\documentclass{article} 
\usepackage[preprint]{colm2026_conference}
\usepackage{tikz}
\usepackage{pgfplots}
\pgfplotsset{compat=1.18}
\usepackage{subcaption}
\usepackage{microtype}
\usepackage{hyperref}
\usepackage{url}
\usepackage{colortbl}
\usepackage{xcolor}
\usepackage{booktabs}
\usepackage{amsmath}
\usepackage{graphicx}
\usepackage{comment}
\usepackage{wrapfig}

\usetikzlibrary{arrows.meta, shapes.misc, shapes.geometric, positioning}
\usepackage{titlesec}
\usepackage{tcolorbox}
  \tcbuselibrary{breakable, skins, listings}
\usepackage{fontawesome5}   

\titlespacing*{\section}{0pt}{0.8ex plus 0.5ex minus 0.2ex}{0.5ex plus 0.2ex}
\titlespacing*{\subsection}{0pt}{0.6ex plus 0.4ex minus 0.2ex}{0.4ex plus 0.2ex}

\usepackage{lineno}

\definecolor{darkblue}{rgb}{0, 0, 0.5}
\hypersetup{colorlinks=true, citecolor=darkblue, linkcolor=darkblue, urlcolor=darkblue}

\title{Multimodal Latent Reasoning via Predictive  Embeddings}

\author{Ashutosh Adhikari \& Mirella Lapata \\
School of Informatics\\
University of Edimburgh\\
\texttt{\{ashutosh.adhikari, mlap\}@ed.ac.uk} \\
}

\begin{document}

\ifcolmsubmission
\linenumbers
\fi

\maketitle

\begin{abstract}
Tool-augmented multimodal reasoning enables visual language models (VLMs) to improve perception by interacting with external tools (e.g.,~cropping, depth estimation). However, such approaches incur substantial inference overhead, require specialized supervision, and are prone to erroneous tool calls. We propose \mbox{\textsc{Pearl}} (\textbf{P}redictive 
\textbf{E}mbedding \textbf{A}lignment for  \textbf{R}easoning in 
\textbf{L}atent space), a JEPA-inspired framework that learns from expert tool-use trajectories entirely in the latent space, eliminating the need for explicit tool invocation at inference time. Unlike reconstruction-based latent reasoning methods, which autoregressively generate latent tokens and suffer from training–inference mismatch and limited support for multi-step tool use, \textsc{Pearl} directly learns predictive embeddings from multimodal trajectories while preserving the standard vision-language generation pipeline: it  is model-agnostic, simple to train, and naturally supports trajectories with multiple tool calls. Experiments across multiple perception benchmarks show that \textsc{Pearl} matches or outperforms standard supervised fine-tuning and reconstruction-based latent reasoning approaches. Furthermore, we provide empirical evidence that reconstruction-based methods primarily learn embeddings rather than image edits in latent space, motivating predictive embedding learning as a more principled alternative.

\end{abstract}

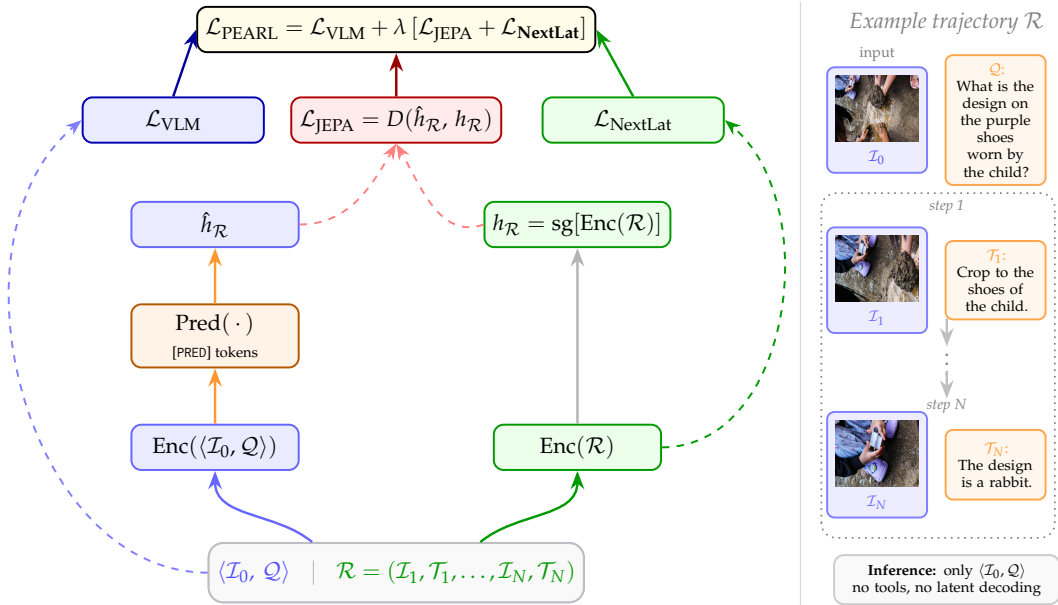
\begin{figure}[t!]
  \centering
  \resizebox{\linewidth}{!}{%
\hspace{-1cm}\begin{tikzpicture}[
  font=\small,
  >=Stealth,
  line width=0.8pt,
  view/.style={
    rectangle, rounded corners=6pt, draw, fill=white,
    minimum width=2.2cm, minimum height=0.75cm,
    align=center, font=\small\itshape
  },
  enc/.style={
    rectangle, rounded corners=4pt, draw=gray!60, fill=gray!10,
    minimum width=2.4cm, minimum height=0.65cm,
    align=center, font=\small
  },
  pred/.style={
    rectangle, rounded corners=4pt, draw=orange!70!black, fill=orange!10,
    minimum width=2.4cm, minimum height=0.65cm,
    align=center, font=\small
  },
  loss/.style={
    rectangle, rounded corners=4pt, draw, fill=white,
    minimum width=2.6cm, minimum height=0.65cm,
    align=center, font=\small\bfseries
  },
  lossjepa/.style={loss, draw=red!70!black,   fill=red!8},
  lossvlm/.style ={loss, draw=blue!70!black,  fill=blue!8},
  lossbs/.style  ={loss, draw=green!60!black, fill=green!8},
  imgbox/.style={
    rectangle, draw=blue!50, fill=blue!7, rounded corners=3pt,
    minimum width=1.3cm, minimum height=1.0cm, align=center, font=\tiny
  },
  txtbox/.style={
    rectangle, draw=orange!70, fill=orange!7, rounded corners=3pt,
    minimum width=1.3cm, minimum height=1.0cm, align=center,
    text width=1.2cm, font=\tiny
  },
  arr/.style={->, gray!70, line width=1pt},
  arrbold/.style={->, line width=1.5pt},
  contrib/.style={->, dashed, line width=0.8pt},
]


\node[
  view, draw=blue!40, fill=white,
  minimum height=0.85cm
] (inputnode) at (2.6, 0)
{%
  \textcolor{blue!70}{$\langle \mathcal{I}_0,\, \mathcal{Q} \rangle$}%
  \quad{\color{gray!40}\rule[-0.3em]{0.5pt}{1.1em}}\quad%
  \textcolor{green!60!black}{$\mathcal{R} = (\mathcal{I}_1,\mathcal{T}_1,\dots,\mathcal{I}_N,\mathcal{T}_N)$}%
};

\node[
  view, draw=gray!50, fill=gray!5,
  minimum height=0.85cm
] (inputnode) at (2.6, 0)
{%
  \textcolor{blue!70}{$\langle \mathcal{I}_0,\, \mathcal{Q} \rangle$}%
  \quad{\color{gray!50}\rule[-0.3em]{0.5pt}{1.1em}}\quad%
  \textcolor{green!60!black}{$\mathcal{R} = (\mathcal{I}_1,\mathcal{T}_1,\dots,\mathcal{I}_N,\mathcal{T}_N)$}%
};

\node[enc, draw=blue!60, fill=blue!8]  (encx) at (0,   1.8) {$\mathrm{Enc}(\langle\mathcal{I}_0,\mathcal{Q}\rangle)$};
\node[enc, draw=green!60!black, fill=green!8] (encr) at (5.2, 1.8) {$\mathrm{Enc}(\mathcal{R})$};

\node[pred] (prednode) at (0, 3.4)
  {$\mathrm{Pred}(\,\cdot\,)$\\[1pt]{\tiny\normalfont[\texttt{PRED}] tokens}};

\node[enc, draw=blue!60, fill=blue!8]
  (hhat) at (0, 5.0) {$\hat{h}_{\mathcal{R}}$};

\node[enc, draw=green!60!black, fill=green!8]
  (hR) at (5.2, 5.0)
  {$h_{\mathcal{R}} = \mathrm{sg}[\mathrm{Enc}(\mathcal{R})]$};

\node[lossvlm]  (Lvlm)  at (-.6,   6.5) {$\mathcal{L}_{\mathrm{VLM}}$};
\node[lossjepa] (Ljepa) at (2.6, 6.5)
  {$\mathcal{L}_{\mathrm{JEPA}} = D(\hat{h}_\mathcal{R},\, h_\mathcal{R})$};
\node[lossbs]   (Lbs)   at (6, 6.5) {$\mathcal{L}_{\mathrm{NextLat}}$};

\node[loss, draw=black, fill=yellow!10, minimum width=4.8cm]
  (Lpearl) at (2.6, 7.8)
  {$\mathcal{L}_{\mathrm{PEARL}} = \mathcal{L}_{\mathrm{VLM}}
    + \lambda\,[\mathcal{L}_{\mathrm{JEPA}} + \mathcal{L}_{\text{NextLat}}]$};


\draw[arr, blue!60]
  (inputnode.160) to[out=140, in=270] (encx.south);
\draw[arr, green!60!black]
  (inputnode.20)  to[out=40,  in=270] (encr.south);

\draw[arr, orange!80]      (encx)     -- (prednode);
\draw[arr, orange!80]      (prednode) -- (hhat);

\draw[arr, gray!60]        (encr)     -- (hR);


\draw[contrib, red!50]
  (hhat.east) to[out=0, in=240] (Ljepa.south);
\draw[contrib, red!50]
  (hR.west)   to[out=200, in=300] (Ljepa.south);

  \draw[contrib, blue!50]
  (inputnode.west) to[out=180, in=220] (Lvlm.west);
  
\draw[contrib, green!60!black]
  (encr.east) to[out=0, in=320] (Lbs.east);

\draw[arr, blue!60!black]  (Lvlm.north)  -- (Lpearl.west);
\draw[arr, red!60!black]   (Ljepa.north) -- (Lpearl.south);
\draw[arr, green!60!black] (Lbs.north)   -- (Lpearl.east);

\draw[gray!30, line width=0.5pt] (8.4, -0.5) -- (8.4, 8.2);


\def\rx{10.5}

\node[font=\footnotesize\itshape, gray] at (\rx, 7.9)
  {Example trajectory $\mathcal{R}$};

\node[imgbox, minimum height=1.1cm] (i0) at (\rx-1.0, 6.5)
  {\includegraphics[width=1.2cm,height=0.95cm]{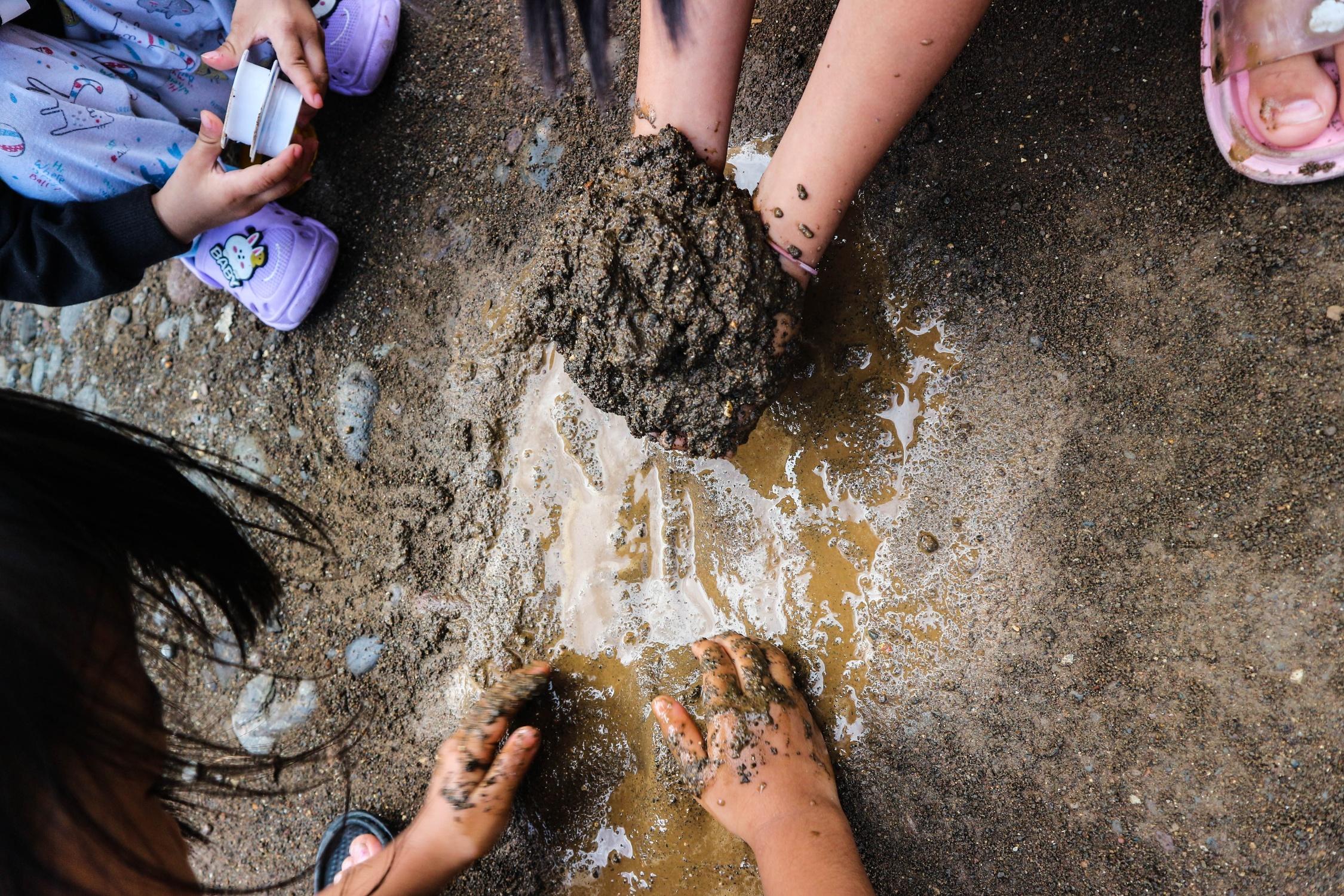}\\[1pt]
   {\color{blue!70}$\mathcal{I}_0$}};
\node[txtbox] (q0) at (\rx+0.7, 6.5)
  {{\color{orange!80}$\mathcal{Q}$:}\\What is the design on the purple shoes worn by the child?};

\node[font=\tiny, gray] at (\rx-1.0, 7.45) {input};
\node[font=\tiny\itshape, gray] at (\rx, 5.25) {step 1};
\node[imgbox] (i1) at (\rx-1.0, 4.2)
  {\includegraphics[width=1.2cm,height=0.95cm]{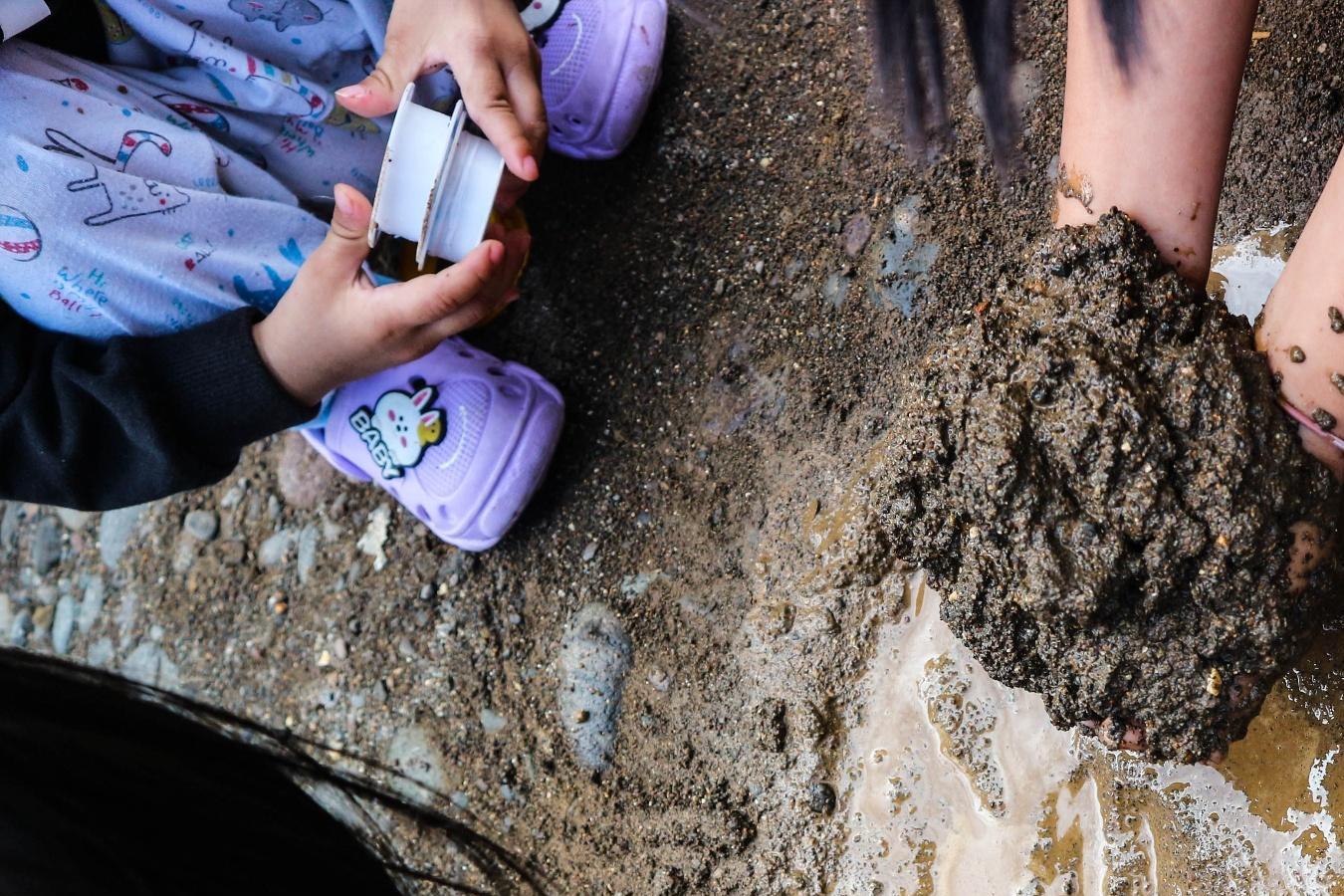}\\[1pt]
   {\color{blue!60}$\mathcal{I}_1$}};
\node[txtbox] (t1) at (\rx+0.7, 4.2)
  {{\color{orange!80}$\mathcal{T}_1$:}\\Crop to the shoes of the child.};

\draw[arr, gray!50] (\rx, 3.65) -- (\rx, 3.25);
\node[font=\normalfont\small, gray] at (\rx, 3.1) {$\vdots$};
\draw[arr, gray!50] (\rx, 2.9) -- (\rx, 2.5);

\node[font=\tiny\itshape, gray] at (\rx, 2.4) {step $N$};
\node[imgbox] (in) at (\rx-1.0, 1.55)
  {\includegraphics[width=1.2cm,height=0.95cm]{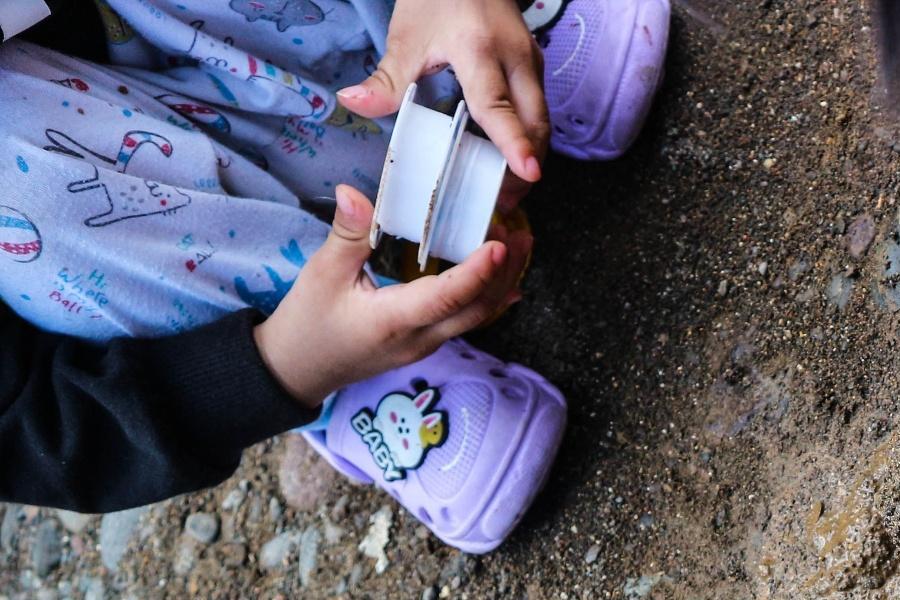}\\[1pt]
   {\color{blue!60}$\mathcal{I}_N$}};
\node[txtbox] (tn) at (\rx+0.7, 1.55)
  {{\color{orange!80}$\mathcal{T}_N$:}\\The design is a rabbit.};

  \draw[gray, rounded corners=6pt, line width=0.7pt, dotted]
  (\rx-1.8, 0.5) rectangle (\rx+1.55, 5.45);

\node[
  rectangle, draw=gray!40, fill=gray!5, rounded corners=4pt,
  text width=3.0cm, align=center, font=\tiny
] at (\rx, -.1)
  {\textbf{Inference:} only $\langle\mathcal{I}_0,\mathcal{Q}\rangle$\\
   no tools,  no latent decoding};

\end{tikzpicture}%
}
\caption{\textbf{\textsc{Pearl} architecture.} \textit{Left:} Solid arrows
  denote forward-pass dataflow; dashed arrows denote which components
  contribute to each loss (no forward pass).  During training, two
  independent forward passes encode
  $\langle\mathcal{I}_0,\mathcal{Q}\rangle$ and the expert trajectory
  $\mathcal{R}$.  A tied-weights predictor maps the input encoding to
  the trajectory latent space.  $\mathcal{L}_{\mathrm{JEPA}}$ aligns
  the predicted embedding $\hat{h}_\mathcal{R}$ with the stop-gradient
  target $h_\mathcal{R}$; $\mathcal{L}_{\mathrm{VLM}}$ preserves
  autoregressive text generation; $\mathcal{L}_{\text{NextLat}}$
  regularises hidden states to act as belief states.  \textit{Right:}
  An example trajectory of sequential visual edits
  $(\mathcal{I}_1,\dots,\mathcal{I}_N)$ interleaved with reasoning
  text $(\mathcal{T}_1,\dots,\mathcal{T}_N)$.
\label{fig:pearl}
}

\vspace{-10pt}
\end{figure}
\section{Introduction}
Recent work in multimodal reasoning has explored augmenting vision-language models (VLMs) with external tools (e.g., cropping, object detection, depth estimation) to improve grounded reasoning \citep{pixel-reasoner, Wu2025VToolR1VL, Su2025OpenThinkIMGLT}. By enabling models to iteratively manipulate visual inputs, these approaches allow VLMs to ``think with images'' rather than relying solely on textual reasoning.

While interacting with expert tools to edit images is an effective strategy for grounding  LLMs in visual context \citep{yue2023mmmu, emma}, tool-based approaches introduce practical and conceptual challenges. First, invoking external tools incurs substantial inference-time latency and compute overhead \citep{nichols2025optimizingagenticlanguagemodel, wu-etal-2025-joint}. Second, learning to correctly select and parameterize tools requires specialized supervision  \citep{Wu2025VToolR1VL, pixel-reasoner}, and even after such training, erroneous calls waste inference compute and pollute the context with irrelevant information. Third,  most approaches assume a homogeneous tool set,  whereas  handling the diverse tools exposed by complex MCP servers requires  advanced planning and  instruction-following \citep{wu-etal-2025-joint}. 

A promising direction replaces explicit tool use with latent
reasoning, where models operate in a continuous embedding space
instead of generating discrete intermediate outputs~\citep{coconut,
  colar, ccot2}. Prior work has explored \emph{reconstruction-based}
latent reasoning, in which models autoregressively generate latent
tokens intended to represent intermediate visual transformations
\citep{li2025latentvisualreasoning,
  yang2025machinementalimageryempower,
  gu2025thinkmorphemergentpropertiesmultimodal}.  Borrowing from
Coconut~\citep{coconut}, these methods supervise continuous latent
tokens with a reconstruction objective against the outputs of visual
tools, while preserving the standard transformer architecture (see
Figure~\ref{fig:lvr_gen} for an overview).  Despite their appeal, these
approaches suffer from two fundamental limitations. First, they
exhibit a training--inference mismatch: during training, models are
supervised with as many latent tokens as there are image patch tokens
in the tool output, yet at inference only a small, fixed number of
latent tokens is decoded, often without improving, and sometimes
degrading, performance~\citep{yang2025machinementalimageryempower,
  li2025latentvisualreasoning}. Second, they are typically confined to
single-step transformations, failing to support multi-step reasoning
over sequences of tool operations.  These observations suggest that
reconstruction-based methods primarily learn useful \emph{embeddings}
rather than genuinely simulating visual transformations in latent
space.

In this work, we propose \textbf{\textsc{Pearl}} (\textbf{P}redictive
\textbf{E}mbedding \textbf{A}lignment for \textbf{R}easoning in
\textbf{L}atent space), a JEPA-inspired framework that learns
predictive representations from expert tool-use trajectories. Rather
than autoregressively generating latent tokens, \textsc{Pearl}
predicts trajectory embeddings from an image–question pair (see
Figure~\ref{fig:pearl}, right), allowing the model to internalize the
effects of tool use without explicit tool invocation. The framework
operates entirely in latent space during training, avoids
training–inference mismatch, and supports multi-step reasoning over
trajectories with multiple tool calls. We instantiate \textsc{Pearl}
by jointly optimizing a standard vision–language generation objective
with a predictive embedding objective over interleaved multimodal
trajectories (see Figure~\ref{fig:pearl}, left), enabling the model to
retain its text generation capabilities while learning both the
\emph{effects} and \emph{sequencing} of task-relevant transformations.

We evaluate \textsc{Pearl} across a range of multimodal reasoning
benchmarks, including settings with single and multiple tool
calls. \textsc{Pearl} consistently matches or outperforms supervised
fine-tuning and reconstruction-based latent reasoning
approaches. Moreover, our analysis shows that reconstruction-based
methods primarily learn embeddings rather than performing genuine
latent ``imagination'', supporting predictive embedding learning as a
more principled alternative.

\section{Related Work}
\textbf{Tool-augmented Multi-modal Reasoning.}
\vspace{-.3cm}
A prominent line of work augments vision-language models with external visual tools to improve grounded reasoning \citep{pixel-reasoner,huang2025visualtoolagentvistareinforcementlearning,zheng2025chainoffocus}. These approaches enable models to iteratively manipulate images through operations such as cropping, object detection, or spatial transformations, effectively allowing them to ``think with images'' rather than relying solely on textual reasoning. More advanced systems further integrate specialized tools such as depth estimation or multi-step visual editing pipelines, often interleaving tool execution with chain-of-thought reasoning. To determine when and how to invoke tools, early methods rely on supervised fine-tuning with expert trajectories \citep{Su2025OpenThinkIMGLT,chung2026v1learningpointvisual} while more recent approaches leverage reinforcement learning to acquire tool-use policies and support multi-step reasoning \citep{zheng2025chainoffocus,pixel-reasoner,geng2026webwatcher}.
Despite their effectiveness, these methods introduce significant practical challenges: tool invocation incurs substantial inference-time  overhead, requires specialized supervision for correct tool selection and parameterization, and remains brittle to errors that can propagate through the reasoning process.  Furthermore, many approaches assume a fixed or homogeneous tool set, limiting scalability to  diverse or dynamically evolving tool environments. We instead explore latent visual reasoning, where the model internalizes the effects of tool use directly in representation space, eliminating the need for explicit tool invocation at inference time.

\textbf{Reconstruction-based Latent Reasoning.}  Existing work on
latent reasoning in multimodal models autoregressively generates
latent tokens under a reconstruction objective, aiming to ``imagine''
intermediate image edits. Concretely, these latent tokens are trained
to reconstruct tool outputs, borrowing from approaches such as
Coconut~\citep{coconut} in the text domain. This requires models to
switch between latent and discrete tokens at inference, a complication
that \textsc{Pearl} avoids. However, this line of work is largely
confined to single tool calls within a reasoning trajectory
\citep{li2025latentvisualreasoning,
  yang2025machinementalimageryempower,
  gu2025thinkmorphemergentpropertiesmultimodal}. COVT~\citep{qin2025covt}
extends this setting to multiple tool calls, but applies a fixed
sequence of operations regardless of the input query. This design
imposes two key limitations. First, COVT avoids dynamic planning by
restricting tool calls to parameter-free operations, precluding
actions such as cropping that require query-specific
arguments. Second, all tool calls are applied directly to the original
image $\mathcal{I}_0$ rather than to the outputs of preceding
operations, resulting in a shallow tree of independent branches rather
than a chain of dependent steps. In contrast, \textsc{Pearl} avoids
autoregressive latent generation by predicting a single embedding of
the full expert trajectory, naturally supporting multi-step tool use
as the prediction target encodes the entire trajectory $\mathcal{R}$
rather than a single transformation. 

\textbf{Joint Embedding Predictive Architectures for Language Models.}
The Joint Embedding Predictive Architecture (JEPA) is a self-supervised learning framework that trains a model to predict the embedding of one view of the data from another, rather than reconstructing raw inputs. It has shown promise as a pre-training objective for multimodal models with  V-JEPA2 \citep{assran2025vjepa2selfsupervisedvideo} and VL-JEPA \citep{chen2026vljepajointembeddingpredictive} achieving competitive performance; however, both require substantial pre-training compute to match the performance of current state-of-the-art models. In the textual domain, LLM-JEPA \citep{huang2025llmjepalargelanguagemodels} adapts 
JEPA  to fine-tune off-the-shelf language models, avoiding the need to train from scratch. Concretely, LLM-JEPA minimises a distance between the encoded representations of paired text and code, encouraging the model to develop modality-agnostic representations conducive to text-to-code generation. Similarly, V-JEPA2 aligns embeddings from video frames and
robotic actions with predicted future frames for video model pre-training.

 \textsc{Pearl} 
 adapts the JEPA objective to fine-tune an off-the-shelf VLM from expert multimodal tool-use trajectories, treating the image-question pair and the full reasoning trajectory as two views of the same problem. This allows \textsc{Pearl} to internalize the effect of sequential visual tool use in the latent space, without incurring the prohibitive compute of pre-training or departing from the standard image-text-to-text inference pipeline.

\section{PEARL: Predictive Latent Reasoning}

\subsection{Problem Formulation and Overview}

We consider a training setting where each example consists of an
image-question pair \(\mathbf{x} = \langle \mathcal{I}_0, \mathcal{Q}
\rangle\) and an expert multimodal reasoning trajectory
\[
\mathcal{R} = (\mathcal{I}_1, \mathcal{T}_1, \mathcal{I}_2, \mathcal{T}_2,
\dots, \mathcal{I}_N, \mathcal{T}_N),
\]
where each \(\mathcal{I}_i\) is an intermediate image produced by an
expert visual tool (e.g., crop, highlight, spatial transformation) and
each \(\mathcal{T}_i\) is the associated reasoning text, with the final
step containing the answer to \(\mathcal{Q}\) (see
Figure~\ref{fig:pearl}).

Our goal is to train a VLM that benefits from such tool-use trajectories
\emph{without} invoking tools at inference time. In contrast to prior
reconstruction-based approaches, which autoregressively generate latent
tokens intended to reconstruct intermediate visual edits
\citep{li2025latentvisualreasoning, yang2025machinementalimageryempower,
gu2025thinkmorphemergentpropertiesmultimodal}, \textsc{Pearl}
(\textbf{P}redictive \textbf{E}mbedding \textbf{A}lignment for
\textbf{R}easoning in \textbf{L}atent space) directly predicts a latent
representation of the full trajectory from the original image-question
pair, preserving the standard VLM inference pipeline while internalizing
information from tool-based reasoning during training.

Concretely, \textsc{Pearl} encodes \(\mathbf{x}\) and \(\mathcal{R}\)
independently and trains a lightweight predictor to anticipate the
trajectory embedding from the input alone.  Intuitively, the predictor asks: given only the image and question,
can the model anticipate what the expert tool-use trajectory would look like
in latent space?  This design has three
advantages. First, it avoids explicit tool invocation at inference.
Second, it avoids the training--inference mismatch of reconstruction-based
methods, where models are trained with many latent tokens but decode only
a small fixed number at test time. Third, because the prediction target
encodes the \emph{entire} multimodal trajectory rather than a single
image edit, \textsc{Pearl} naturally supports multiple tool calls.
\subsection{Trajectory Encoding}

We instantiate both $\mathrm{Enc}(\mathbf{x})$ and $\mathrm{Enc}(\mathcal{R})$
using the hidden states of an off-the-shelf autoregressive VLM, serializing the two
views as follows (see also Figure~\ref{fig:pearl}):
\begin{itemize}
    \item the \textbf{input view} consists of the original image-question pair
          $\langle \mathcal{I}_0, \mathcal{Q} \rangle$;
    \item the \textbf{trajectory view} consists of the interleaved sequence
          $(\mathcal{I}_1, \mathcal{T}_1, \ldots, \mathcal{I}_N, \mathcal{T}_N)$.
\end{itemize}
For each view, we run a forward pass through the VLM and take the final hidden
state of the last token as its sequence representation, following prior work on
JEPA-style fine-tuning of decoder-only language
models~\citep{huang2025llmjepalargelanguagemodels}:
\[
h_{\mathbf{x}} = \mathrm{Enc}(\mathbf{x}), \qquad
h_{\mathcal{R}} = \mathrm{Enc}(\mathcal{R}).
\]

Using separate forward passes avoids cross-view information leakage and keeps
the method architecture-agnostic, at the cost of additional training-time
compute. This overhead applies only during training; inference remains identical
to standard VLM decoding. A lightweight predictor network then takes
the input encoding~$h_\mathbf{x}$ and produces a predicted version of
the  trajectory embedding.

\subsection{Latent Trajectory Predictor}

To map the input representation \(h_{\mathbf{x}}\) to the trajectory
latent space, we use a predictor built from the VLM itself. Following
prior tied-weights JEPA formulations
\citep{huang2025llmjepalargelanguagemodels}, we append \(K\) learnable
special tokens \texttt{[PRED]} to the serialized input \(\mathbf{x}\),
and define the predicted trajectory representation as the hidden state
of the final predictor token:
\[
\hat{h}_{\mathcal{R}} = \mathrm{Pred}(h_{\mathbf{x}}).
\]

Intuitively, the predictor allows the model to perform additional
nonlinear computation over the image-question representation before
producing the target latent (see visualization in Figure~\ref{fig:pearl}). When \(K=0\), the predictor reduces to
the identity map. In practice, using predictor tokens lets us reuse
the VLM's existing self-attention stack rather than introducing a
separate MLP or auxiliary transformer, thereby keeping the method
simple and parameter-efficient.

\subsection{Predictive Embedding Objective}

Our central training signal is a JEPA-style predictive embedding loss that aligns
the predicted latent \(\hat{h}_{\mathcal{R}}\) with the encoded expert trajectory
\(h_{\mathcal{R}}\). We define
\begin{equation}
\mathcal{L}_{\rm JEPA}
=
D\!\left(
\hat{h}_{\mathcal{R}},
\mathrm{sg}[h_{\mathcal{R}}]
\right),
\label{eq:jepa}
\end{equation}
where \(D(\cdot,\cdot)\) is a distance function and \(\mathrm{sg}[\cdot]\) denotes
stop-gradient. In our experiments, we use SmoothL1 loss for \(D\).

This objective encourages the model to learn a compact representation of the
\emph{effect} of expert tool use and multimodal reasoning, rather than explicitly
reconstructing intermediate image edits. In this sense, PEARL learns predictive
trajectory embeddings rather than latent image generation (see Figure~\ref{fig:pearl}).

\subsection{Next-Latent Prediction}

A potential limitation of using the final hidden state of a decoder as
a sequence representation is that it may not reliably summarize all
relevant preceding context.  To encourage hidden states to behave as
predictive summary states, we add a next-latent prediction objective
inspired by recent work on latent dynamics in transformers
\citep{teoh2025nextlatentpredictiontransformerslearn}.

Let \(h_t\) denote the hidden state at time step \(t\) within the
serialized trajectory. A lightweight latent predictor is trained to
forecast future hidden states \(\hat{h}_{t+i}\) from the current state
\(h_t\), for a prediction horizon \(d\). We optimize
\begin{equation}
\mathcal{L}_{\text{NextLat}} =
\mathbb{E}_{t}
\left[
\frac{1}{d}
\sum_{i=1}^{d}
\mathrm{SmoothL1Loss}
\left(
\mathrm{sg}[h_{t+i}],
\hat{h}_{t+i}
\right)
\right].
\label{eq:bs_rewrite}
\end{equation}

This objective encourages hidden states to be informative about future trajectory evolution, making them better suited for sequence-level latent alignment. Specifically, \citet{teoh2025nextlatentpredictiontransformerslearn} show that optimizing the hidden state transitions as in Equation~(\ref{eq:bs_rewrite}) causes them to converge to \textit{belief states}, which \citet{kaelblingplanning} define as sufficient statistics of the past history. We view this term as a regularizer that improves the quality of the learned latent representations, rather than as a separate reasoning mechanism.

\subsection{Autoregressive Generation Objective}

In addition to latent alignment, we retain the standard VLM training
objective over the textual portions of the expert
trajectory. Given the interleaved multimodal context, the
model is trained to autoregressively predict each token in the textual
segments \(\mathcal{T}_1, \dots, \mathcal{T}_N\):
\begin{equation}
\mathcal{L}_{\rm VLM}
=
-
\sum_{n=1}^{N}
\sum_{t=1}^{|\mathcal{T}_n|}
\log p_\theta
\!\left(
\mathcal{T}_n^{(t)}
\mid
\mathcal{I}_0, \mathcal{Q},
\mathcal{I}_1, \mathcal{T}_1,
\dots,
\mathcal{I}_n, \mathcal{T}_n^{(<t)}
\right).
\label{eq:vlm_rewrite}
\end{equation}
Here $\mathcal{T}_n^{(t)}$ denotes the $t$-th token of the $n$-th
reasoning step, $\mathcal{T}_n^{(<t)}$ denotes all preceding tokens
within that step, and the conditioning context includes all prior
image-text pairs $(\mathcal{I}_1, \mathcal{T}_1, \dots, \mathcal{I}_{n-1},
\mathcal{T}_{n-1})$ as well as the original input
$\langle \mathcal{I}_0, \mathcal{Q} \rangle$. This term ensures that
PEARL preserves the VLM's standard text generation capability, which
is necessary for producing final answers at test time.

\subsection{Training Objective}

We jointly optimize the autoregressive generation objective, the predictive
embedding objective, and the next-latent regularizer:
\begin{equation}
\mathcal{L}_{\rm PEARL}
=
\mathcal{L}_{\rm VLM}
+
\lambda\,[\mathcal{L}_{\rm JEPA}
+
\mathcal{L}_{\text{NextLat}}],
\label{eq:pearl_final}
\end{equation}
where $\lambda$ jointly controls the contribution of both latent
objectives relative to the generation loss, reflecting the view that
$\mathcal{L}_{\rm JEPA}$ and $\mathcal{L}_{\text{NextLat}}$ together
constitute a single latent learning signal (see Figure~\ref{fig:pearl}).

These three terms play complementary roles. \(\mathcal{L}_{\rm VLM}\)
preserves the model's ability to generate answers in the discrete
token space.  \(\mathcal{L}_{\rm JEPA}\) teaches the model to predict
a latent representation of expert multimodal reasoning from the
original image-question pair.  
$\mathcal{L}_{\text{NextLat}}$ encourages hidden states to act as
belief states (i.e., sufficient summaries of past context), making them
more informative encoding targets for $\mathcal{L}_{\rm JEPA}$.

\subsection{Inference}

At inference time, PEARL requires only the original image-question
pair \(\langle \mathcal{I}_0, \mathcal{Q} \rangle\), and answers using
the standard generation pipeline of the underlying VLM. It does
\emph{not} invoke external tools, does \emph{not} generate
intermediate edited images, does \emph{not} use \texttt{[PRED]}
tokens, and does \emph{not} autoregressively decode latent reasoning
tokens. The cost of learning from tool use is shifted entirely to
training time, preserving simple and efficient inference.

\section{Experimental Setting}

\textbf{Training Regimes.}
To demonstrate the effectiveness of \textsc{Pearl} at learning from
expert tool-use trajectories, we finetune models across three settings:
(i)~single-type, single tool call per trajectory;
(ii)~multiple-type, single tool call per trajectory; and
(iii)~single-type, multiple tool calls per trajectory.
We leave the multiple-type, multiple tool call setting to future work,
as no open-source training data currently exists for this combination.

For setting~(i), we use the data from LVR~\citep{li2025latentvisualreasoning},
which provides regions of interest used to crop the input image
$\mathcal{I}_0$, forming the trajectory $\mathcal{R}$.
For setting~(ii), we use the ThinkMorph
dataset~\citep{gu2025thinkmorphemergentpropertiesmultimodal}, which
contains four equal-sized subsets corresponding to different tool types:
bounding boxes over regions of interest, highlights over charts, jigsaw
puzzle reconstructions, and spatial navigation paths over maze images.
For setting~(iii), we use the PixelReasoner
dataset~\citep{pixel-reasoner}, where each trajectory $\mathcal{R}$
contains up to three sequential crops of $\mathcal{I}_0$. Dataset
statistics and examples are provided in
Appendix~\ref{sec:dataset-examples}. 

\textbf{Evaluation Benchmarks.}
Following previous work (e.g., \citealt{li2025latentvisualreasoning})
we evaluate \textsc{Pearl} on a suite of perception intensive visual
question answering (VQA; \citealt{antol2015vqa}) benchmarks.  These
include V*~\citep{vstar}, which tests models' ability to perform
visual search for objects and their attributes (V*$_{DA}$) and to
identify relative positions of objects (V*$_{RP}$). We further
evaluate on five subsets of the Blink benchmark~\citep{Blink}:
Counting, IQ, Jigsaw, Spatial Relation, and Relative
Reflectance. Finally, we include MMVP~\citep{MMVP}, which probes
perceptual robustness using image pairs that CLIP treats as similar
despite clear visual differences. All benchmarks are formulated as
multiple-choice tasks, enabling straightforward answer parsing (see
Appendix~\ref{sec:dataset-examples} for details).


\textbf{Comparison Models.}
Our primary comparisons use Qwen2.5-VL-7B-Instruct \citep{qwen2.5-VL}, enabling direct head-to-head
evaluation against all reconstruction-based baselines. To further demonstrate \textsc{Pearl}'s model-agnostic nature, we also report
results for the smaller Qwen2.5-VL-3B-Instruct variant and the 4B variant of Qwen3-VL~\citep{bai2025qwen3vltechnicalreport}.

For the single-type, single tool call setting, we compare against
LVR~\citep{li2025latentvisualreasoning}, using their released
HuggingFace checkpoint, with 4~latent tokens (aka 4 steps) which the
authors note yields the best overall quality. We also compare
against CoVT~\citep{qin2025covt}, reporting results directly from the
original paper. LVR achieves the strongest performance among
reconstruction-based latent reasoning methods (see
Figure~\ref{fig:lvr_gen} for an illustration). 

For the multiple-type, single tool call setting, we compare
\textsc{Pearl} against a LoRA-finetuned variant trained on the
ThinkMorph data~\citep{gu2025thinkmorphemergentpropertiesmultimodal}.
We do not compare against the original ThinkMorph model, as it relies
on explicit intermediate image generation at inference, making it
incomparable with latent reasoning methods.\footnote{We were also
unable to reproduce the original ThinkMorph results, as the released
checkpoint and evaluation scripts were not available in a complete
form at the time of submission.} For the single-type, multiple tool
call setting, we compare directly against
PixelReasoner's~(\citeyear{pixel-reasoner}) released model, which
invokes tools explicitly at inference.

\textsc{Pearl} is finetuned with LoRA~\citep{llora:2021}  adapters (rank $r=64$ and $\alpha=128$). 
Across  settings, we include a
LoRA SFT baseline trained on the same data as \textsc{Pearl} and the
instruction-tuned model without fine-tuning as a zero-shot
baseline. Hyperparameter settings for \textsc{Pearl} are provided in
Appendix~\ref{sec:appendix}.

\section{Results}
\label{sec:results}

\subsection{How Does \textsc{Pearl} Compare to Reconstruction-based Methods?}

\definecolor{sectiongray}{RGB}{210,210,210}

\begin{table}[t]
\centering
\setlength{\tabcolsep}{5pt}
\renewcommand{\arraystretch}{1.15}
\resizebox{\textwidth}{!}{%
\begin{tabular}{lcccccccccc}
\toprule
\textcolor{black}{\textbf{Model}}
  & \textcolor{black}{\textbf{V$^{*}$}}
  & \textcolor{black}{\textbf{V$^{*}_{DA}$}}
  & \textcolor{black}{\textbf{V$^{*}_{RP}$}}
  & \textcolor{black}{\textbf{MMVP}}
  & \textcolor{black}{\textbf{Counting}}
  & \textcolor{black}{\textbf{IQ}}
  & \textcolor{black}{\textbf{Jigsaw}}
  & \textcolor{black}{\textbf{Rel.\ Ref}}
  & \textcolor{black}{\textbf{Spatial Rel}} \\
\midrule

\rowcolor{sectiongray}
\multicolumn{10}{c}{\textit{\small No fine-tuning}} \\
Qwen2.5-VL-7B-Instruct
  & 78.5 & 81.7 & 73.7 & 66.7 & 66.7 & 26.0 & 52.0 & 38.8 & 87.4 \\

\midrule
\rowcolor{sectiongray}
\multicolumn{10}{c}{\textit{\small Single-type, single tool call}} \\
{CoVT}~\citep{qin2025covt}
  & 78.0 & --- & --- & 58.7 & --- & --- & --- & --- & --- \\
LVR~\citep{li2025latentvisualreasoning} (4 steps)
  & 80.1 & 85.2 & 73.7 & 72.0 & 68.3 & 26.0 & 51.3 & \textbf{41.0} & 89.5 \\
SFT (LVR data)
  & 79.1 & 82.6 & 73.7 & 65.7 & 67.5 & 26.7 & 45.3 & 33.6 & 88.8 \\
\textsc{Pearl} (LVR data)
  & \textbf{81.5} & \textbf{86.1} & \textbf{74.5} & \textbf{73.5}
  & 68.3 & \textbf{28.2} & \textbf{53.1} & 39.6 & 89.5 \\

\midrule
\rowcolor{sectiongray}
\multicolumn{10}{c}{\textit{\small Multiple-type, single tool call}} \\
SFT (ThinkMorph data)
  & 42.4 & 58.3 & 18.4 & 36.7 & 38.3 & 16.7 & 22.0 & 38.8 & 60.1 \\
\textsc{Pearl} (ThinkMorph data)
  & \textbf{73.8} & \textbf{76.5} & \textbf{69.7} & \textbf{75.3}
  & \textbf{65.0} & \textbf{26.0} & \textbf{53.3} & \textbf{46.3} & \textbf{88.8} \\

\midrule
\rowcolor{sectiongray}
\multicolumn{10}{c}{\textit{\small Single-type, multiple tool calls}} \\
PixelReasoner~\citep{pixel-reasoner}
  & \textbf{80.1} & \textbf{81.7} & \textbf{77.6} & 67.0
  & 66.7 & 25.3 & 52.7 & \textbf{42.5} & 88.1 \\
\textsc{Pearl} (PixelReasoner data)
  & 79.1 & \textbf{81.7} & 75.0 & \textbf{70.0}
  & \textbf{70.0} & \textbf{28.7} & \textbf{53.3} & 40.3 & \textbf{89.5} \\

\bottomrule
\end{tabular}%
}
\caption{Results for Qwen2.5-VL-7B-Instruct across all training settings.
         \textsc{Pearl} requires no
         tool calls at inference time. \textbf{Bold} denotes best result
         per block.}
\label{tab:main}
\end{table}
Table~\ref{tab:main} compares \textsc{Pearl} against various
Qwen2.5-VL-7B-Instruct baselines and comparison systems  (across three training settings).
As can be seen, \textsc{Pearl} consistently matches or outperforms
its respective baselines while requiring no tool calls at inference, 
an advantage none of the reconstruction-based or tool-augmented methods
share.

\textbf{Single-type, single tool call.}
\textsc{Pearl} trained on LVR data outperforms both the SFT baseline
and LVR~\citep{li2025latentvisualreasoning} (4~steps) on V$^*$ (81.5
vs.\ 79.1 and 80.1) and MMVP (73.5 vs.\ 65.7 and 72.0), while matching
LVR on Spatial Rel (89.5) and improving on Jigsaw (53.1 vs.\ 51.3).
Notably, LVR finetunes the entire decoder whereas \textsc{Pearl} uses
only LoRA adapters, making these gains more parameter-efficient.
CoVT~\citep{qin2025covt} underperforms even the zero-shot baseline on
MMVP (58.7 vs.\ 66.7), suggesting its fixed-sequence design is poorly
suited to this benchmark.

\textbf{Multiple-type, single tool call.}
The ThinkMorph results are the most striking in the table.
\textsc{Pearl} outperforms the SFT baseline by over 31 points on V$^*$
(73.8 vs.\ 42.4) and more than doubles it on MMVP (75.3 vs.\ 36.7).
The SFT baseline collapses under the heterogeneity of four qualitatively
different tool types, whereas \textsc{Pearl}'s trajectory-level embedding
target is agnostic to tool type, explaining its robustness across the
full ThinkMorph benchmark.

\textbf{Single-type, multiple tool calls.}
\textsc{Pearl} is competitive with PixelReasoner~\citep{pixel-reasoner},
which explicitly invokes tools at inference time. \textsc{Pearl}
outperforms it on MMVP (70.0 vs.\ 67.0), Counting (70.0 vs.\ 66.7),
Jigsaw (53.3 vs.\ 52.7), IQ (28.7 vs.\ 25.3), and Spatial Rel (89.5
vs.\ 88.1), while PixelReasoner leads on V$^*_{RP}$ (77.6 vs.\ 75.0)
and Rel.\ Ref (42.5 vs.\ 40.3). The fact that \textsc{Pearl} matches an
inference-time tool-use system while operating as a standard
image-to-text model demonstrates that the tool-use signal can be
effectively internalized during training through predictive embedding
alignment.

Ablations in Appendix~\ref{sec:appendix} confirm that encouraging hidden states to act as belief states
meaningfully improves the quality of the learned trajectory embeddings.  Figure~\ref{fig:vizplot} provides further support: 
t-SNE visualizations show that \textsc{Pearl} induces coherent clusters that
align the two views ($\langle \mathcal{I}_0, \mathcal{Q} \rangle$ and
$\mathcal{R}$) across tasks, whereas SFT produces fragmented clusters,
confirming that predictive embedding alignment learns more structured
representations.

\subsection{What is the Effect of Training Regime on \textsc{Pearl}?}

Although
no single training regime dominates uniformly across all benchmarks,
clear patterns emerge. The LVR regime (single-type, single tool
call) is strongest on \emph{visual search} tasks, yielding the highest scores
on V$^*$ (81.5) and V$^*_{DA}$ (86.1). The PixelReasoner regime
(single-type, multiple tool calls) performs best on tasks requiring
\emph{counting and spatial reasoning}, leading on Counting (70.0) and matching
the best result on Spatial Rel (89.5). The ThinkMorph regime
(multiple-type, single tool call) stands out on \emph{perceptual robustness}
benchmarks, leading clearly on MMVP (75.3) and Rel.\ Ref (46.3),
suggesting that exposure to diverse tool types improves fine-grained
perceptual discrimination.
Taken together, these results indicate that the three regimes are
complementary rather than competing.
A natural direction for future work is a combined training strategy
that draws on all three regimes simultaneously, which we would expect
to yield stronger across-the-board performance.
\begin{table}[t]
\centering
\setlength{\tabcolsep}{5pt}
\renewcommand{\arraystretch}{1.15}
\resizebox{\textwidth}{!}{%
\begin{tabular}{lcccccccccc}
\toprule
\textcolor{black}{\textbf{Model}}
  & \textcolor{black}{\textbf{V$^{*}$}}
  & \textcolor{black}{\textbf{V$^{*}_{DA}$}}
  & \textcolor{black}{\textbf{V$^{*}_{RP}$}}
  & \textcolor{black}{\textbf{MMVP}}
  & \textcolor{black}{\textbf{Counting}}
  & \textcolor{black}{\textbf{IQ}}
  & \textcolor{black}{\textbf{Jigsaw}}
  & \textcolor{black}{\textbf{Rel.\ Ref}}
  & \textcolor{black}{\textbf{Spatial Rel}} \\
\midrule

\rowcolor{sectiongray}
\multicolumn{10}{c}{\textit{\small Qwen2.5-VL-3B-Instruct}} \\
No fine-tuning & 56.0 & 53.0 & 60.5 & 59.3 & 65.8 & 26.0 & 45.3 & 44.8 & 80.4 \\ 
LVR~\citep{li2025latentvisualreasoning} (4 steps)
  & 64.9 & 69.6 & 60.5 & 54.7 & --- & 29.3 & 52.7 & --- & --- \\
\textsc{Pearl} (LVR data)
  & \textbf{73.8} & \textbf{82.6} & 60.5 & \textbf{68.7}
  & \textbf{66.7} & 29.3 & 51.3 & 41.8 & \textbf{84.6} \\
\textsc{Pearl} (PixelReasoner data)
  & 69.6 & 78.3 & 56.6 & 63.7
  & 67.5 & \textbf{30.7} & 49.3 & \textbf{43.3} & 81.8 \\
\textsc{Pearl} (ThinkMorph data) & 62.8 & 69.6 & 52.6 & 60.0 & 61.7 & 29.3 & 43.3 & 35.1 & 78.3 \\
\midrule
\rowcolor{sectiongray}
\multicolumn{10}{c}{\textit{\small Qwen3-VL-4B-Instruct}} \\
No fine-tuning
  & 81.2 & 86.1 & 73.7 & 75.7 & 65.8 & 24.0 & 68.0 & 62.7 & 83.9 \\
   \textsc{Pearl} (LVR data) & 81.7 & 85.2 & 76.3 & 80.0 & 67.5 & 27.3 & 70.0 & 53.7 & 85.3\\
\textsc{Pearl} (PixelReasoner data)
  & 79.6 & 82.6 & \textbf{75.0} & \textbf{77.3} & \textbf{70.8} & \textbf{25.3} & 68.0 & 57.5 & \textbf{87.4} \\
  \textsc{Pearl} (ThinkMorph data) & 75.4 & 81.7 & 65.8 & 76.3 & 66.7& 26.7 & 75.3 & 66.4 & 81.8 \\

\bottomrule
\end{tabular}%
}
\caption{Results for smaller model variants (Qwen2.5-VL-3B-Instruct and Qwen3-VL-4B-Instruct) across three training regimes. 
         \textbf{Bold} denotes best result per block.
}
\label{tab:small_models}
\end{table}

\subsection{Does \textsc{Pearl} Generalise Across Model Sizes and Architectures?}

Table~\ref{tab:small_models} reports results for smaller model variants,
demonstrating that \textsc{Pearl}'s gains are not specific to the 7B
scale or to the Qwen2.5 architecture.

\textsc{Pearl} trained on LVR data substantially outperforms the 3B LVR
baseline across nearly all benchmarks --- most strikingly on V$^*$ (73.8
vs.\ 64.9) and MMVP (68.7 vs.\ 54.7) --- despite using only LoRA
adapters. This mirrors the pattern observed at 7B and confirms that
predictive embedding learning scales down gracefully. The PixelReasoner-trained
variant performs slightly lower overall but remains competitive on
Counting and Rel.\ Ref, consistent with the regime-specific patterns
observed in Table~\ref{tab:main}.

The zero-shot Qwen3-VL-4B baseline is already strong, particularly on
Jigsaw (68.0) and Rel.\ Ref (62.7), which substantially exceed the
corresponding 7B zero-shot scores, reflecting the stronger 
perceptual capabilities of the Qwen3 architecture. \textsc{Pearl} trained
on LVR data improves further on V$^*$ (81.7 vs.\ 81.2) and MMVP (80.0
vs.\ 75.7), while the PixelReasoner-trained variant gains on
Spatial Rel (87.4 vs.\ 83.9). Both variants show some regression on
Rel.\ Ref relative to the zero-shot baseline, which we leave to future
investigation.


For the 3B variant, \textsc{Pearl} trained on ThinkMorph underperforms
the LVR-trained variant across most benchmarks (V$^*$: 62.8 vs.\ 73.8;
MMVP: 60.0 vs.\ 68.7). This is likely due to two compounding factors:
the diverse tool-type signal may require greater model capacity, and
ThinkMorph's verbose, open-ended reasoning steps introduce a
training--test format mismatch that smaller models struggle to overcome
when producing multiple-choice answers. By contrast, the 4B Qwen3-VL
variant benefits strongly from the ThinkMorph regime, achieving 75.3
on Jigsaw (vs.\ 68.0 zero-shot) and 66.4 on Rel.\ Ref
(vs.\ 62.7 zero-shot), surpassing both the LVR- and PixelReasoner-trained
variants on these benchmarks — consistent with the 7B finding that
diverse tool exposure improves perceptual discrimination.

Across both a smaller and a newer model architecture, \textsc{Pearl}
consistently matches or improves over its respective fine-tuning
baselines without any architecture-specific modifications, confirming
its model-agnostic nature. The 
 complementary strengths of the three training regimes identified at 7B (visual search (LVR), spatial and counting abilities (PixelReasoner), and
perceptual robustness (ThinkMorph)) generalise across scales, although the
ThinkMorph gains appear sensitive to base model capacity.


\subsection{Do Reconstruction-based Methods Actually ``Imagine'' Images?}

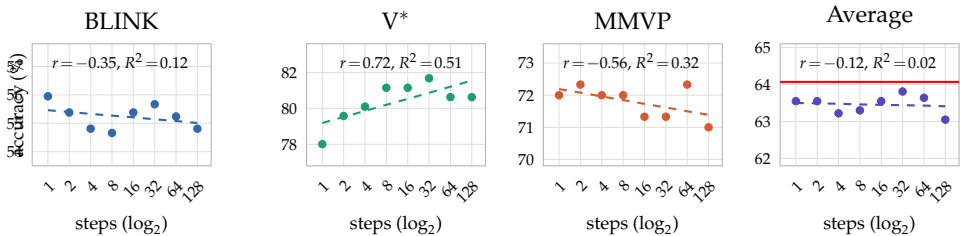
\begin{figure*}[t]
\begin{center}

\begin{subfigure}[b]{0.27\textwidth}
\begin{tikzpicture}
\begin{axis}[
width=1.05\textwidth, height=0.85\textwidth,
  xmode=log, log basis x=2,
  xmin=0.6, xmax=200,
  xtick={1,2,4,8,16,32,64,128},
  xticklabels={1,2,4,8,16,32,64,128},
  x tick label style={font=\tiny, rotate=45, anchor=north east},
  y tick label style={font=\tiny},
  xlabel style={yshift=4pt},
  xlabel={\scriptsize steps ($\log_2$)},
  ylabel={\scriptsize accuracy (\%)},
  ylabel style={yshift=-15pt},
  title style={yshift=-4pt},
title={\footnotesize BLINK},
  ymin=53.5, ymax=57.8,
  grid=both,
  grid style={line width=0.3pt, gray!25},
  tick style={draw=none},
  axis line style={gray!60},
]
  \addplot[only marks, mark=*, mark size=1.4pt,
           color={rgb,255:red,50;green,102;blue,173}]
    coordinates {
      (1,  55.95)(2,  55.38)(4,  54.81)(8,  54.66)
      (16, 55.38)(32, 55.67)(64, 55.24)(128,54.81)
    };
  \addplot[dashed, thick, color={rgb,255:red,50;green,102;blue,173}]
    coordinates {
      (1,  55.46)(2,  55.40)(4,  55.33)(8,  55.27)
      (16, 55.21)(32, 55.14)(64, 55.08)(128,55.01)
    };
  \node[anchor=north east, font=\tiny, text=black]
    at (axis cs:128,57.8) {$r\!=\!-0.35$, $R^2\!=\!0.12$};
\end{axis}
\end{tikzpicture}
\end{subfigure}
\hspace{-.3cm}
\begin{subfigure}[b]{0.27\textwidth}
\begin{tikzpicture}
\begin{axis}[
width=1.05\textwidth, height=0.85\textwidth,
  xmode=log, log basis x=2,
  xmin=0.6, xmax=200,
  xtick={1,2,4,8,16,32,64,128},
  xticklabels={1,2,4,8,16,32,64,128},
  x tick label style={font=\tiny, rotate=45, anchor=north east},
  y tick label style={font=\tiny},
  xlabel style={yshift=4pt},
  xlabel={\scriptsize steps ($\log_2$)},
  title style={yshift=-4pt},
title={\footnotesize V$^{*}$},
  ymin=76.8, ymax=83.6,
  grid=both,
  grid style={line width=0.3pt, gray!25},
  tick style={draw=none},
  axis line style={gray!60},
]
  \addplot[only marks, mark=*, mark size=1.4pt,
           color={rgb,255:red,29;green,158;blue,117}]
    coordinates {
      (1,  78.01)(2,  79.58)(4,  80.10)(8,  81.15)
      (16, 81.15)(32, 81.68)(64, 80.63)(128,80.63)
    };
  \addplot[dashed, thick, color={rgb,255:red,29;green,158;blue,117}]
    coordinates {
      (1,  79.19)(2,  79.52)(4,  79.86)(8,  80.20)
      (16, 80.53)(32, 80.87)(64, 81.21)(128,81.55)
    };
  \node[anchor=north east, font=\tiny, text=black]
    at (axis cs:128,83.6) {$r\!=\!0.72$, $R^2\!=\!0.51$};
\end{axis}
\end{tikzpicture}
\end{subfigure}
\hspace{-.8cm}
\begin{subfigure}[b]{0.27\textwidth}
\begin{tikzpicture}
\begin{axis}[
width=1.05\textwidth, height=0.85\textwidth,
  xmode=log, log basis x=2,
  xmin=0.6, xmax=200,
  xtick={1,2,4,8,16,32,64,128},
  xticklabels={1,2,4,8,16,32,64,128},
  x tick label style={font=\tiny, rotate=45, anchor=north east},
  y tick label style={font=\tiny},
  xlabel style={yshift=4pt},
  xlabel={\scriptsize steps ($\log_2$)},
  title style={yshift=-4pt},
title={\footnotesize MMVP},
  ymin=69.8, ymax=73.6,
  grid=both,
  grid style={line width=0.3pt, gray!25},
  tick style={draw=none},
  axis line style={gray!60},
]
  \addplot[only marks, mark=*, mark size=1.4pt,
           color={rgb,255:red,216;green,90;blue,48}]
    coordinates {
      (1,  72.00)(2,  72.33)(4,  72.00)(8,  72.00)
      (16, 71.33)(32, 71.33)(64, 72.33)(128,71.00)
    };
  \addplot[dashed, thick, color={rgb,255:red,216;green,90;blue,48}]
    coordinates {
      (1,  72.19)(2,  72.08)(4,  71.96)(8,  71.85)
      (16, 71.73)(32, 71.62)(64, 71.50)(128,71.39)
    };
  \node[anchor=north east, font=\tiny, text=black]
    at (axis cs:128,73.6) {$r\!=\!-0.56$, $R^2\!=\!0.32$};
\end{axis}
\end{tikzpicture}
\end{subfigure}
\hspace{-.8cm}
\begin{subfigure}[b]{0.27\textwidth}
\begin{tikzpicture}
\begin{axis}[
width=1.05\textwidth, height=0.85\textwidth,
  xmode=log, log basis x=2,
  xmin=0.6, xmax=200,
  xtick={1,2,4,8,16,32,64,128},
  xticklabels={1,2,4,8,16,32,64,128},
  x tick label style={font=\tiny, rotate=45, anchor=north east},
  y tick label style={font=\tiny},
  xlabel style={yshift=4pt},
  xlabel={\scriptsize steps ($\log_2$)},
  title style={yshift=-4pt},
title={\footnotesize Average},
  ymin=61.8, ymax=65.1,
  grid=both,
  grid style={line width=0.3pt, gray!25},
  tick style={draw=none},
  axis line style={gray!60},
]
  \addplot[only marks, mark=*, mark size=1.4pt,
           color={rgb,255:red,83;green,74;blue,183}]
    coordinates {
      (1,  63.55)(2,  63.55)(4,  63.22)(8,  63.30)
      (16, 63.55)(32, 63.81)(64, 63.64)(128,63.05)
    };
  \addplot[dashed, thick, color={rgb,255:red,83;green,74;blue,183}]
    coordinates {
      (1,  63.50)(2,  63.49)(4,  63.48)(8,  63.46)
      (16, 63.45)(32, 63.44)(64, 63.43)(128,63.41)
    };
  \addplot[solid, thick, red]
    coordinates {(0.6,64.07)(200,64.07)};
  \node[anchor=north east, font=\tiny, text=black]
    at (axis cs:128,65.1) {$r\!=\!-0.12$, $R^2\!=\!0.02$};
\end{axis}
\end{tikzpicture}
\end{subfigure}
\end{center}
\vspace{-12pt}
\caption{Correlation between the number of reasoning steps and accuracy across
BLINK ($n{=}697$), V$^{*}$($n{=}191$), MMVP ($n{=}300$), and on
average ($n{=}1{,}188$). The x-axis uses a $\log_2$ scale; dashed lines show
the log-linear trend. Near-zero $r$ and $R^2$ values confirm that embedding
quality is stable across reasoning steps. The red line is \textsc{Pearl} trained on LVR.}
\label{fig:step-correlation}
\end{figure*}

A central motivation for \textsc{Pearl} is the observation that
reconstruction-based latent reasoning methods
\citep{li2025latentvisualreasoning,
  yang2025machinementalimageryempower,qin2025covt} may not be doing
what they claim.  These methods assume that autoregressively
generating latent tokens allows a model to ``imagine'' intermediate
image edits in the latent space, and that more tokens should
correspond to a more complete imagined transformation.
\begin{wrapfigure}[16]{r}[0pt]{0.45\linewidth}
    \vspace{-5pt}
    \includegraphics[width=\linewidth]{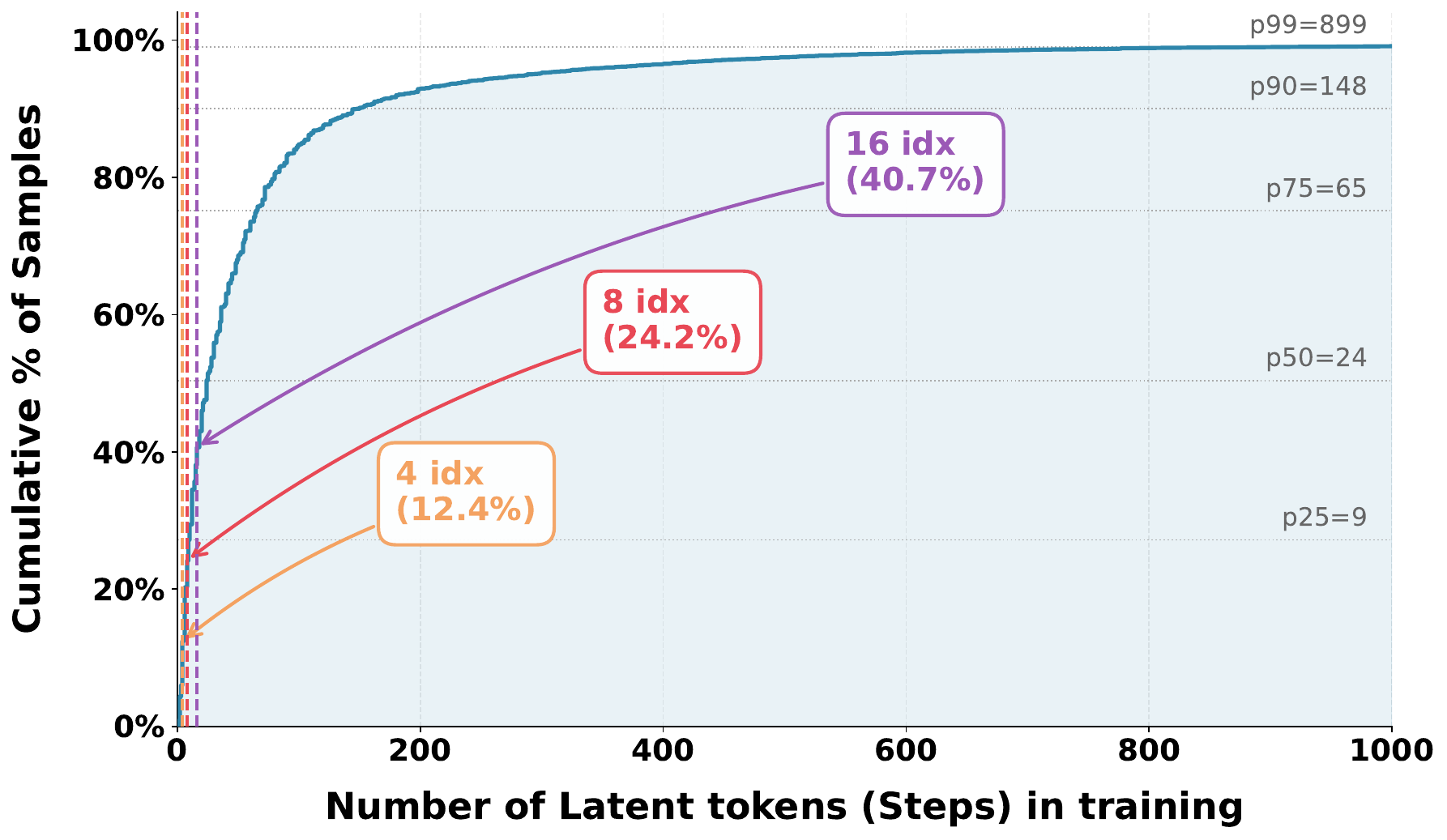}
    \caption{Cumulative distribution function (CDF) of the number of
      latent tokens per training example (x-axis) over a sample of
      $\sim$19k examples used to train LVR.}
    \label{fig:lvrdat_statistics}
\end{wrapfigure}
Figure~\ref{fig:lvrdat_statistics} shows that over 75\% of the edited
images used to supervise LVR contain more than 8 latent tokens during
training, as a direct consequence of the token count scaling with the
number of image patch tokens in each example, yet LVR fixes this to
just 4 or 8 tokens at inference.  Figure~\ref{fig:step-correlation}
further reveals that model quality does not improve as the number of
latent tokens increases, and in some cases slightly degrades, with a
weakly negative correlation across BLINK and MMVP. In fact, using just
1 or 2 latent tokens achieves parity with much higher token counts.
This training--inference mismatch, combined with the insensitivity of
performance to token count, suggests that reconstruction-based methods
are not genuinely simulating visual transformations in latent
space. Instead, they appear to learn useful \emph{embeddings}: compact
representations that improve answer quality regardless of how many
latent tokens are decoded. This finding directly motivates
\textsc{Pearl}: if reconstruction-based methods are learning
embeddings anyway, it is more principled to learn these
directly via a predictive objective, without the added complexity of
autoregressive latent generation and  the practical burden of switching
between continuous and discrete tokens at inference.

\section{Conclusion}
We presented \textsc{Pearl}, a JEPA-inspired framework that learns
from expert tool-use trajectories in the latent space without
requiring explicit tool invocation at inference. Rather than
reconstructing intermediate image edits autoregressively,
\textsc{Pearl} directly predicts a trajectory-level embedding from the
image-question pair, preserving the standard VLM inference
pipeline. Across three training regimes and multiple perception
benchmarks, \textsc{Pearl} consistently matches or outperforms
reconstruction-based methods and SFT baselines using only LoRA
adapters, with gains that generalise across model sizes and
architectures. Our analysis further challenges the premise of
reconstruction-based latent reasoning: performance is largely
insensitive to the number of latent tokens decoded at inference,
suggesting these methods learn useful embeddings rather than genuinely
imagining image edits. A natural direction for future work is a
combined training strategy that draws on all three regimes
simultaneously, as well as extending \textsc{Pearl} to settings with
diverse, multi-step tool use and explicit latent planning at inference
(see Appendix~\ref{sec:limitations} for  further discussion). 

\section*{Ethics Statement}

This work presents \textsc{Pearl}, a framework for training
vision-language models to internalize the effects of visual tool use
in the latent space. We discuss the ethical considerations most
relevant to this research.

\textbf{Intended Use and Misuse.}
\textsc{Pearl} is designed to improve the efficiency and accuracy of
multimodal reasoning in VLMs for perception-intensive tasks. As with
any method that improves the capability of language models, there is
potential for misuse in applications that generate misleading visual
interpretations or automate harmful decision-making. We encourage
practitioners to apply appropriate safeguards when deploying systems
built on this work in high-stakes settings.

\textbf{Data and Bias.}
Our experiments rely on publicly available datasets (LVR, ThinkMorph,
PixelReasoner) and pre-trained models (Qwen2.5-VL, Qwen3-VL). Any
biases present in these data sources or base models may be inherited
or amplified by \textsc{Pearl}. We did not conduct a systematic bias
audit and caution against deployment in sensitive domains without
further evaluation.

\textbf{Environmental Cost.}
Training \textsc{Pearl} requires two forward passes per example,
roughly doubling compute relative to standard fine-tuning. All
experiments were conducted on H100 and H200 GPUs. We partially
mitigate this cost by using LoRA adapters rather than full fine-tuning,
and by training for a limited number of steps or epochs per regime.

\textbf{Broader Impact.}
By eliminating the need for explicit tool invocation at inference,
\textsc{Pearl} reduces the latency and resource cost of deploying
tool-augmented VLMs, which may make capable multimodal reasoning
more accessible. We release model weights and code to support
reproducibility and further research.

\bibliography{colm2026_conference}
\bibliographystyle{colm2026_conference}
\newpage
\appendix
\section{Additional Results and Hyperparameter Settings}
\label{sec:appendix}

\subsection{Hyperparameter Settings}
\label{ref:hyperparams}
We set $\lambda$ in Equation~(\ref{eq:pearl_final}) to $0.2$ and the
number of \texttt{[PRED]} tokens to $4$ across all training settings.

All experiments are conducted on either 4 NVIDIA H200 or 6 NVIDIA H100
GPUs. We train on LVR data for 2{,}500 steps, on PixelReasoner data for
4 epochs, and on ThinkMorph data for 1 epoch, selecting the best
checkpoint based on validation loss. On H200s, we use a per-device
batch size of 4 with gradient accumulation of 4; on H100s, we reduce
the per-device batch size to 2 to fit within memory. For all runs, LoRA
adapters are configured with rank $r=64$ and $\alpha=128$.

At inference, we constrain model outputs to the option letter using a
maximum of 4 tokens, enabling straightforward answer parsing and
ensuring fair, direct comparison with LVR~\citep{li2025latentvisualreasoning}.

\subsection{Embedding Visualisation}
\label{sec:app_embedding_viz}
\begin{figure}[t]
\includegraphics[width=\linewidth]{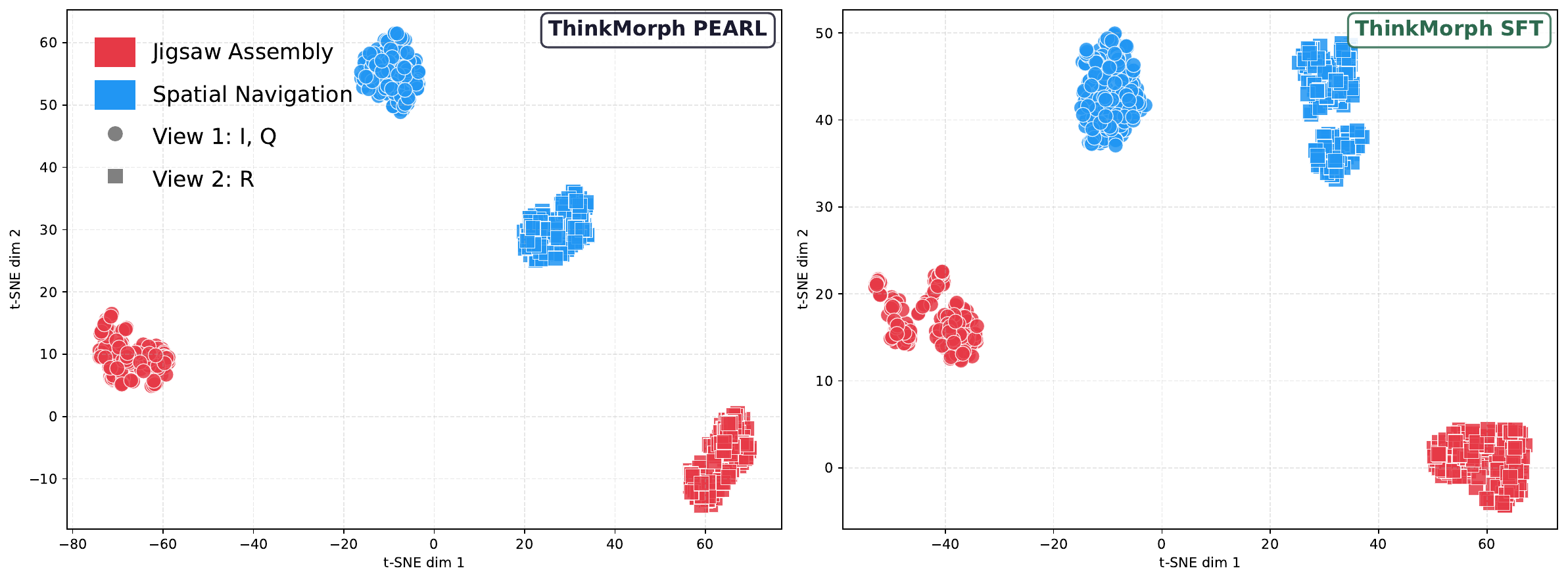}
    \caption{T-SNE visualization of the views $I_0, Q$ and $R$ across tasks for Qwen2.5-VL-7B-Instruct trained with \textsc{Pearl} on the left, compared with simple fine-tuning with next-token prediction on the right.}
    \label{fig:vizplot}
\end{figure}

Figure~\ref{fig:vizplot} visualizes t-SNE projections of the embeddings
learned by \textsc{Pearl} and a LoRA SFT baseline, both trained on
ThinkMorph data. For \textsc{Pearl}, the two views, i.e., 
$\langle \mathcal{I}_0, \mathcal{Q} \rangle$ and $\mathcal{R}$,  form
coherent, well-separated clusters that align across tasks (Jigsaw and
Spatial Navigation), indicating that the predictive embedding objective
encourages the model to develop shared, task-discriminative
representations of the input and trajectory. By contrast, the SFT
baseline produces fragmented clusters in which the two views are not
consistently aligned, suggesting that next-token prediction alone does
not induce the same degree of structured latent organisation. This
qualitative difference is consistent with \textsc{Pearl}'s quantitative
gains and supports the view that the JEPA objective encourages more
semantically meaningful representations than standard fine-tuning.

\subsection{Ablation for Next-Latent Prediction}
\label{sec:ablation_next_lat}

In this section, we show that using next latent predictions for training the hidden states $h_t$ to be the \textit{belief states} for representing views, aids model quality.

\begin{table}[t]
\centering
\setlength{\tabcolsep}{5pt}
\renewcommand{\arraystretch}{1.15}
\resizebox{\textwidth}{!}{%
\begin{tabular}{lcccccccccc}
\toprule
\textcolor{black}{\textbf{Model}}
  & \textcolor{black}{\textbf{V$^{*}$}}
  & \textcolor{black}{\textbf{V$^{*}_{DA}$}}
  & \textcolor{black}{\textbf{V$^{*}_{RP}$}}
  & \textcolor{black}{\textbf{MMVP}}
  & \textcolor{black}{\textbf{Counting}}
  & \textcolor{black}{\textbf{IQ}}
  & \textcolor{black}{\textbf{Jigsaw}}
  & \textcolor{black}{\textbf{Rel.\ Ref}}
  & \textcolor{black}{\textbf{Spatial Rel}} \\
\midrule
\rowcolor{sectiongray}
\multicolumn{10}{c}{\textit{\small Single-type, single tool call}} \\
\textsc{Pearl}
  & \textbf{81.5} & \textbf{86.1} & 74.5 & \textbf{73.5}
  & 68.3 & \textbf{28.2} & \textbf{53.1} & 39.6 & \textbf{89.5} \\
\midrule
\textsc{Pearl} w/o $\mathcal{L}_{\text{NextLat}}$
  & 80.1 & 83.5 & \textbf{75.0} & 69.3
  & 65.0 & 24.0 & 52.7 & 42.5 & 89.5 \\
\textsc{Pearl} w/o $\mathcal{L}_{\text{NextLat}}, \mathcal{L}_{\text{JEPA}}$
  & 79.1 & 82.6 & 73.7 & 65.7
  & 67.5 & 26.7 & 45.3 & 33.6 & 88.8 \\
\bottomrule
\end{tabular}%
}
\caption{Ablation of $\mathcal{L}_{\text{NextLat}}$ on the single-type,
single tool call setting (LVR data, Qwen2.5-VL-7B-Instruct). Removing
the next-latent prediction objective consistently degrades performance,
confirming that encouraging hidden states to act as belief states
improves the trajectory embeddings learned by $\mathcal{L}_{\text{JEPA}}$.
\textbf{Bold} denotes best result per column.}
\label{tab:ablation_next_lat}
\end{table}

Table~\ref{tab:main} ablates the contribution of $\mathcal{L}_{\text{NextLat}}$
by comparing \textsc{Pearl} against a variant trained without this objective.
Removing $\mathcal{L}_{\text{NextLat}}$ leads to consistent degradation across
most benchmarks, with the most notable drops on V$^*$ (80.1 vs.\ 81.5),
V$^*_{DA}$ (83.5 vs.\ 86.1), MMVP (69.3 vs.\ 73.5), and IQ (24.0
vs.\ 28.2). The only benchmark where the ablated variant is competitive
is V$^*_{RP}$ (75.0 vs.\ 74.5), suggesting that the next-latent objective
is most beneficial for tasks requiring holistic visual understanding
rather than simple relative positioning.

These results support the theoretical motivation for $\mathcal{L}_{\text{NextLat}}$:
by encouraging hidden states to converge to belief states (sufficient
summaries of past context),  the objective produces more informative
encoding targets for $\mathcal{L}_{\text{JEPA}}$. Without this regularizer,
the final hidden state of the decoder is a less reliable sequence
representation, which in turn weakens the predictive embedding alignment
signal. $\mathcal{L}_{\text{NextLat}}$ therefore acts as a necessary
complement to $\mathcal{L}_{\text{JEPA}}$ rather than a redundant auxiliary
objective.

\section{Examples and Dataset Statistics}
\label{sec:dataset-examples}

\usetikzlibrary{
  positioning,
  arrows.meta,
  calc,
  fit,
  backgrounds,
  decorations.pathreplacing,
  decorations.pathmorphing,
  shapes.geometric,
  patterns
}
\definecolor{imgtoken}{RGB}{230,126,34}    
\definecolor{texttoken}{RGB}{142,68,173}   
\definecolor{latenttoken}{RGB}{41,128,185} 
\definecolor{specialtoken}{RGB}{127,140,141}
\definecolor{toolcolor}{RGB}{39,174,96}    
\definecolor{lossred}{RGB}{231,76,60}      
\definecolor{bgbox}{RGB}{245,248,250}      
\definecolor{sceneblue}{RGB}{100,149,237}
\definecolor{depthbrown}{RGB}{139,90,43}
\begin{figure}[t]
\begin{tikzpicture}[
    >=Stealth,
    token/.style={
      minimum width=0.38cm,
      minimum height=0.55cm,
      inner sep=0pt,
      draw=black!30,
      line width=0.4pt,
      rounded corners=1.5pt
    },
    imgtok/.style={token, fill=imgtoken!70},
    txttok/.style={token, fill=texttoken!50},
    lattok/.style={token, fill=latenttoken!55},
    spectok/.style={token, fill=specialtoken!40},
    bigblock/.style={
      draw=black!60,
      fill=bgbox,
      rounded corners=4pt,
      minimum height=1.1cm,
      line width=0.8pt,
      font=\sffamily\small\bfseries,
      align=center
    },
    toolblock/.style={
      draw=toolcolor!80,
      fill=toolcolor!12,
      rounded corners=4pt,
      minimum height=0.9cm,
      line width=0.8pt,
      font=\sffamily\small\bfseries,
      align=center
    },
    labelstyle/.style={
      font=\sffamily\scriptsize,
      text=black!70,
      align=center
    },
    arrowline/.style={
      -{Stealth[length=4pt,width=3pt]},
      line width=0.6pt,
      black!60
    },
    lossarrow/.style={
      -{Stealth[length=4pt,width=3pt]},
      line width=0.8pt,
      lossred!80,
      dashed
    },
    wavyarrow/.style={
      -{Stealth[length=4pt,width=3pt]},
      line width=0.7pt,
      latenttoken!80,
      decorate,
      decoration={snake, amplitude=1.2pt, segment length=5pt, pre length=3pt, post length=4pt}
    },
    bracelab/.style={
      font=\sffamily\scriptsize,
      text=black!80
    }
  ]
 
  \node[draw=black!40, rounded corners=3pt, minimum width=1.6cm, minimum height=1.6cm,
        fill=black!5, line width=0.6pt]
    (inputimg) at (0, 0) {};
  \fill[sceneblue!30, rounded corners=2pt] (-0.77, -0.77) rectangle (0.77, 0.77);
  \fill[green!30!black!40] (-0.77, -0.77) rectangle (0.77, -0.3);
  \fill[orange!60] (-0.25, -0.3) rectangle (0.25, 0.15);
  \fill[lossred!70] (-0.35, 0.15) -- (0, 0.45) -- (0.35, 0.15) -- cycle;
  \fill[yellow!80] (0.45, 0.45) circle (0.12);
  \node[font=\sffamily\tiny, align=center, text=black!70] at (0, 1.1) {Input Image $\mathbf{x}_{\mathrm{img}}$};
 
  \node[bigblock, minimum width=12cm] (transformer) at (7, 0)
    {Transformer Backbone};
 
  \foreach \i in {0,...,3} {
    \node[imgtok] (itok\i) at (2.2+\i*0.5, -1.8) {};
  }
  \node[font=\scriptsize, text=black!50] at (4.5, -1.8) {$\cdots$};
  \foreach \i in {0,...,3} {
    \node[txttok] (qtok\i) at (5.1+\i*0.5, -1.8) {};
  }
  \node[spectok, font=\sffamily\tiny] (starttok) at (7.3, -1.8) {};
  \node[font=\tiny, text=black!50] at (7.3, -2.4) {\texttt{<lat>}};
  \foreach \i in {0,...,4} {
    \node[lattok] (ltok\i) at (8.0+\i*0.5, -1.8) {};
  }
  \node[spectok, font=\sffamily\tiny] (endtok) at (10.7, -1.8) {};
  \node[font=\tiny, text=black!50] at (10.7, -2.4) {\texttt{</lat>}};
  \foreach \i in {0,...,2} {
    \node[txttok] (atok\i) at (11.4+\i*0.5, -1.8) {};
  }
 
  \draw[decorate, decoration={brace, mirror, amplitude=4pt}, black!50]
    ([yshift=-4pt]itok0.south west) -- ([yshift=-4pt]itok3.south east)
    node[midway, below=10pt, bracelab] {image embeddings};
  \draw[decorate, decoration={brace, mirror, amplitude=4pt}, black!50]
    ([yshift=-4pt]qtok0.south west) -- ([yshift=-4pt]qtok3.south east)
    node[midway, below=10pt, bracelab] {text query};
  \draw[decorate, decoration={brace, mirror, amplitude=4pt}, latenttoken!70]
    ([yshift=-4pt]ltok0.south west) -- ([yshift=-4pt]ltok4.south east)
    node[midway, below=10pt, bracelab, text=latenttoken!90] {latent reasoning tokens};
  \draw[decorate, decoration={brace, mirror, amplitude=4pt}, black!50]
    ([yshift=-4pt]atok0.south west) -- ([yshift=-4pt]atok2.south east)
    node[midway, below=10pt, bracelab] {text answer};
 
  \foreach \i in {0,...,3} {
    \draw[arrowline] (itok\i.north) -- ++(0, 0.85);
  }
  \foreach \i in {0,...,3} {
    \draw[arrowline] (qtok\i.north) -- ++(0, 0.85);
  }
  \draw[arrowline] (starttok.north) -- ++(0, 0.85);
  \foreach \i in {0,...,4} {
    \draw[arrowline] (ltok\i.north) -- ++(0, 0.85);
  }
  \draw[arrowline] (endtok.north) -- ++(0, 0.85);
  \foreach \i in {0,...,2} {
    \draw[arrowline] (atok\i.north) -- ++(0, 0.85);
  }
 
  \foreach \i in {0,...,4} {
    \node[lattok] (olat\i) at (8.0+\i*0.5, 1.8) {};
  }
  \foreach \i in {0,...,2} {
    \node[txttok] (oatok\i) at (11.4+\i*0.5, 1.8) {};
  }
 
  \foreach \i in {0,...,4} {
    \draw[arrowline] ([yshift=2pt]olat\i.south) ++(0,-0.85) -- (olat\i.south);
  }
  \foreach \i in {0,...,2} {
    \draw[arrowline] ([yshift=2pt]oatok\i.south) ++(0,-0.85) -- (oatok\i.south);
  }

 
  \node[draw=toolcolor!50, rounded corners=3pt, minimum width=1.6cm, minimum height=1.6cm,
        fill=toolcolor!8, line width=0.6pt]
    (toolout) at (4.5, 3.9) {};
  \fill[depthbrown!80, rounded corners=1pt] (3.73, 3.13) rectangle (5.27, 3.43);
  \fill[depthbrown!50] (3.73, 3.43) rectangle (5.27, 3.73);
  \fill[depthbrown!25] (3.73, 3.73) rectangle (5.27, 4.03);
  \fill[depthbrown!10] (3.73, 4.03) rectangle (5.27, 4.33);
  \draw[draw=toolcolor!50, rounded corners=3pt, line width=0.6pt]
    (3.73, 3.13) rectangle (5.27, 4.67);
  \node[font=\sffamily\tiny, align=center, text=toolcolor!90]
    at (4.5, 5.0) {Tool Output $\mathbf{z}_{\mathrm{tool}}$\\[-2pt]{\tiny (depth / segm.)}};
 
  \node[draw=lossred!60, fill=lossred!10, rounded corners=3pt,
        minimum width=1.6cm, minimum height=0.55cm,
        font=\sffamily\scriptsize\bfseries, text=lossred!90]
    (loss_latent) at ($ (olat0)!0.5!(olat4) + (0, 2.1) $ ) {$\mathcal{L}_{\mathrm{latent}}$};
 
  \draw[lossarrow, toolcolor!80] (toolout.east) -- (loss_latent.west)
    node[midway, above, font=\sffamily\tiny, text=toolcolor!80] {supervise};
 
  \draw[lossarrow] ($ (olat0)!0.5!(olat4) + (0, 0.3) $) -- (loss_latent.south);
 
  \node[font=\sffamily\tiny, text=lossred!80, align=center,
        above=0.35cm of loss_latent] {continuous supervision};
 
  \node[draw=lossred!60, fill=lossred!10, rounded corners=3pt,
        minimum width=1.6cm, minimum height=0.55cm,
        font=\sffamily\scriptsize\bfseries, text=lossred!90]
    (loss_text) at (11.9, 3.9) {$\mathcal{L}_{\mathrm{text}}$};
 
  \draw[lossarrow] (11.9, 2.1) -- (loss_text.south);
 
  \node[font=\sffamily\tiny, text=lossred!80, align=center,
        above=0.35cm of loss_text] {discrete supervision};
 
  \node[draw=texttoken!40, fill=texttoken!8, rounded corners=2pt,
        minimum width=1.3cm, minimum height=0.5cm,
        font=\sffamily\tiny, align=center, text=texttoken!90]
    (gttext) at (11.9, 5.8) {Ground-Truth\\[-1pt]Text $\mathbf{y}_{\mathrm{text}}$};
 
  \draw[lossarrow, texttoken!60] (gttext.south) -- (loss_text.north);
 
  \node[draw=imgtoken!60, fill=imgtoken!12, rounded corners=3pt,
        minimum width=1.6cm, minimum height=0.7cm,
        font=\sffamily\tiny\bfseries, align=center]
    (imgenc) at (0, -1.8) {Image\\Encoder};
 
  \draw[arrowline, imgtoken!70] (inputimg.south) -- (imgenc.north);
  \draw[arrowline, imgtoken!70] (imgenc.east) -- (itok0.west);
 
  \draw[decorate, decoration={brace, amplitude=4pt}, latenttoken!70]
    ([yshift=4pt]olat0.north west) -- ([yshift=4pt]olat4.north east)
    node[midway, above=10pt, bracelab, text=latenttoken!90] {predicted latent states};
 
  \draw[decorate, decoration={brace, amplitude=4pt}, texttoken!70]
    ([yshift=4pt]oatok0.north west) -- ([yshift=4pt]oatok2.north east)
    node[midway, above=10pt, bracelab, text=texttoken!90] {predicted text};
 
  \node[font=\scriptsize, text=black!50] at (7.05, -1.8) {$\cdots$};
  \node[font=\scriptsize, text=black!50] at (10.35, -1.8) {$\cdots$};
 
  \begin{scope}[shift={(1.3, -4.5)}]
    \node[font=\sffamily\scriptsize\bfseries, anchor=west] at (0, 0) {Legend:};
    \node[imgtok, minimum width=0.35cm, minimum height=0.35cm] at (1.5, 0) {};
    \node[font=\sffamily\tiny, anchor=west] at (1.75, 0) {Image Embedding};
    \node[txttok, minimum width=0.35cm, minimum height=0.35cm] at (4.0, 0) {};
    \node[font=\sffamily\tiny, anchor=west] at (4.25, 0) {Text Token};
    \node[lattok, minimum width=0.35cm, minimum height=0.35cm] at (5.9, 0) {};
    \node[font=\sffamily\tiny, anchor=west] at (6.15, 0) {Latent Token (continuous)};
    \node[spectok, minimum width=0.35cm, minimum height=0.35cm] at (9.0, 0) {};
    \node[font=\sffamily\tiny, anchor=west] at (9.25, 0) {Special Delimiter};
  \end{scope}
 
\end{tikzpicture}
  \caption{\label{fig:lvr_gen}
    \textbf{Training architecture for latent-augmented multimodal reasoning.}
    The input image $\mathbf{x}_{\mathrm{img}}$ is encoded into image embeddings and
    concatenated with text query tokens. The model autoregressively generates a sequence
    of continuous \emph{latent reasoning tokens} (delimited by \texttt{<lat>} \ldots
    \texttt{</lat>}), followed by discrete text answer tokens.
    Latent tokens are supervised with a continuous regression loss
    $\mathcal{L}_{\mathrm{latent}}$ against a visual tool output
    $\mathbf{z}_{\mathrm{tool}}$ (e.g.\ a depth map or segmentation mask),
    while text tokens are supervised with a standard cross-entropy loss
    $\mathcal{L}_{\mathrm{text}}$ against ground-truth text $\mathbf{y}_{\mathrm{text}}$.
  }
\end{figure}
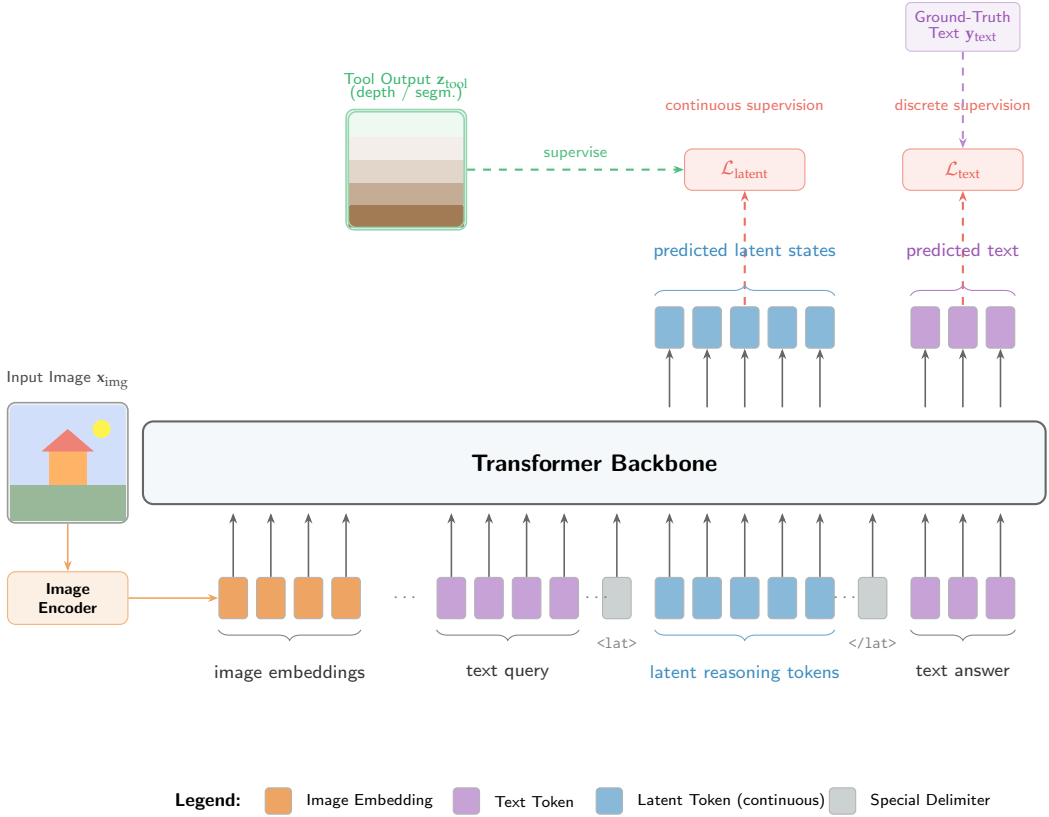

\subsection{Reconstruction-based Multimodal Latent Reasoning}
\label{app:genlvr}

Figure~\ref{fig:lvr_gen} illustrates the general training architecture
shared by reconstruction-based latent reasoning methods. The input
image is first passed through an image encoder to produce a sequence
of image embeddings, which are concatenated with text query tokens and
fed into a transformer backbone. The model is then trained to
autoregressively generate a sequence of continuous latent reasoning
tokens, delimited by special \texttt{<lat>}~$\ldots$~\texttt{</lat>}
markers, before switching to discrete text generation to produce the
final answer. The latent tokens are supervised with a continuous
regression loss $\mathcal{L}_{\mathrm{latent}}$  against
the output of an external visual tool (e.g., a depth map or
segmentation mask), while the text answer tokens are supervised with a
standard cross-entropy loss $\mathcal{L}_{\mathrm{text}}$
against ground-truth text. A key characteristic of this design is the
autoregressive dependency among latent tokens: each generated latent
token is fed back as input to predict the next, effectively requiring
the model to ``imagine'' the tool output token-by-token before
transitioning back to discrete generation. This training-inference
asymmetry --- where many latent tokens are used during training but
only a small fixed number are decoded at test time --- is a central
limitation that \textsc{Pearl} is designed to avoid.


\subsection{Training Data}
The LVR dataset \citep{li2025latentvisualreasoning} contains over 450k training samples; however, we find
that loss curves plateau well before exhausting the data, and therefore
train for at most 2{,}500 steps, reporting results on the best
checkpoint. The ThinkMorph dataset \citep{gu2025thinkmorphemergentpropertiesmultimodal} comprises 6k samples per subset
across four tool types, yielding 24k training samples in total. Its
reasoning steps are notably more verbose than those in LVR or
PixelReasoner, which contributes to the training--test distribution
mismatch observed for smaller models in Section~\ref{sec:results}. We utilize the SFT dataset from PixelReasoner \citep{pixel-reasoner}, which contains $\sim4k$ samples with sequential image transformations over $\mathcal{I}_0$, comprising between 0 and 3 sequential crops of the original image. 

Below, we provide examples of training samples from Thinkmorph
\citep{gu2025thinkmorphemergentpropertiesmultimodal}, Viscot
\citep{shao2024visual}--which was used by LVR
\citep{li2025latentvisualreasoning}, and PixelReasoner's
\citep{pixel-reasoner} finetuning dataset.


\definecolor{SystemBg}{HTML}{F3F0FF}      
\definecolor{SystemBorder}{HTML}{BDB4E8}
\definecolor{UserBg}{HTML}{EFF6FF}        
\definecolor{UserBorder}{HTML}{93C5FD}
\definecolor{AssistantBg}{HTML}{F0FDF4}   
\definecolor{AssistantBorder}{HTML}{86EFAC}
\definecolor{ToolCallBg}{HTML}{FFFBEB}    
\definecolor{ToolCallBorder}{HTML}{FCD34D}
\definecolor{BoxFrame}{HTML}{6366F1}      
\definecolor{BoxTitle}{HTML}{FFFFFF}
\definecolor{LabelSystem}{HTML}{7C3AED}
\definecolor{LabelUser}{HTML}{1D4ED8}
\definecolor{LabelAssistant}{HTML}{15803D}
\definecolor{LabelTool}{HTML}{B45309}

\tcbset{
  datasetbox/.style={
    enhanced, breakable,
    colframe=BoxFrame, colback=white,
    arc=4pt, outer arc=4pt,
    boxrule=1.5pt,
    left=6pt, right=6pt, top=4pt, bottom=6pt,
    attach boxed title to top left={yshift=-2mm, xshift=8pt},
    boxed title style={
      colback=BoxFrame, colframe=BoxFrame,
      arc=3pt, boxrule=0pt,
      left=6pt, right=6pt, top=2pt, bottom=2pt
    },
    fonttitle=\bfseries\small\color{BoxTitle},
  }
}

\newenvironment{rolebubble}[4]{%
  \begin{tcolorbox}[
    enhanced, breakable,
    colback=#1, colframe=#2,
    arc=3pt, boxrule=0.6pt,
    left=7pt, right=7pt, top=4pt, bottom=4pt,
    before upper={\textbf{\textcolor{#3}{\footnotesize #4}}\par\smallskip},
  ]%
}{%
  \end{tcolorbox}\smallskip
}

\newenvironment{toolcall}{%
  \begin{tcolorbox}[
    enhanced,
    colback=ToolCallBg, colframe=ToolCallBorder,
    arc=2pt, boxrule=0.5pt,
    left=5pt, right=5pt, top=2pt, bottom=2pt,
    before upper={\texttt{\textbf{\textcolor{LabelTool}{\scriptsize \faWrench\ tool\_call}}}\par\smallskip},
    fontupper=\ttfamily\scriptsize,
  ]%
}{%
  \end{tcolorbox}%
}


\begin{tcolorbox}[datasetbox, attach boxed title to top center={yshift=-2mm},
  title={PixelReasoner Example \#13: Visual Question Answering with Tool Use}]

\begin{rolebubble}{SystemBg}{SystemBorder}{LabelSystem}{System}
You are a helpful assistant. You may call one or more functions to assist with the
user query. Available tools:
\begin{itemize}
  \item \texttt{crop\_image\_normalized(bbox\_2d, target\_image)} — zoom into a
        bounding-box region of an image.
  \item \texttt{select\_frames(target\_frames)} — select frames from a video.
\end{itemize}
For each function call, return a JSON object inside
\texttt{<tool\_call>...</tool\_call>} tags.
\end{rolebubble}

\begin{rolebubble}{UserBg}{UserBorder}{LabelUser}{User}
\textbf{What does the text on the pink lantern say?}

\medskip
\begin{tabular}{@{}ll@{}}
  \textbf{A:} Sakura Festival & \textbf{B:} Cherry Blossom Viewing \\
  \textbf{C:} Hanami Celebration & \textbf{D:} Spring Lanterns
\end{tabular}

\medskip\noindent
\textit{\footnotesize Guidelines: Understand the visual information and the query.
Determine whether to use the provided tools. Reason step by step and put your
final answer inside \texttt{\{\}.}}

\medskip
\begin{center}
  \includegraphics[width=0.65\linewidth]{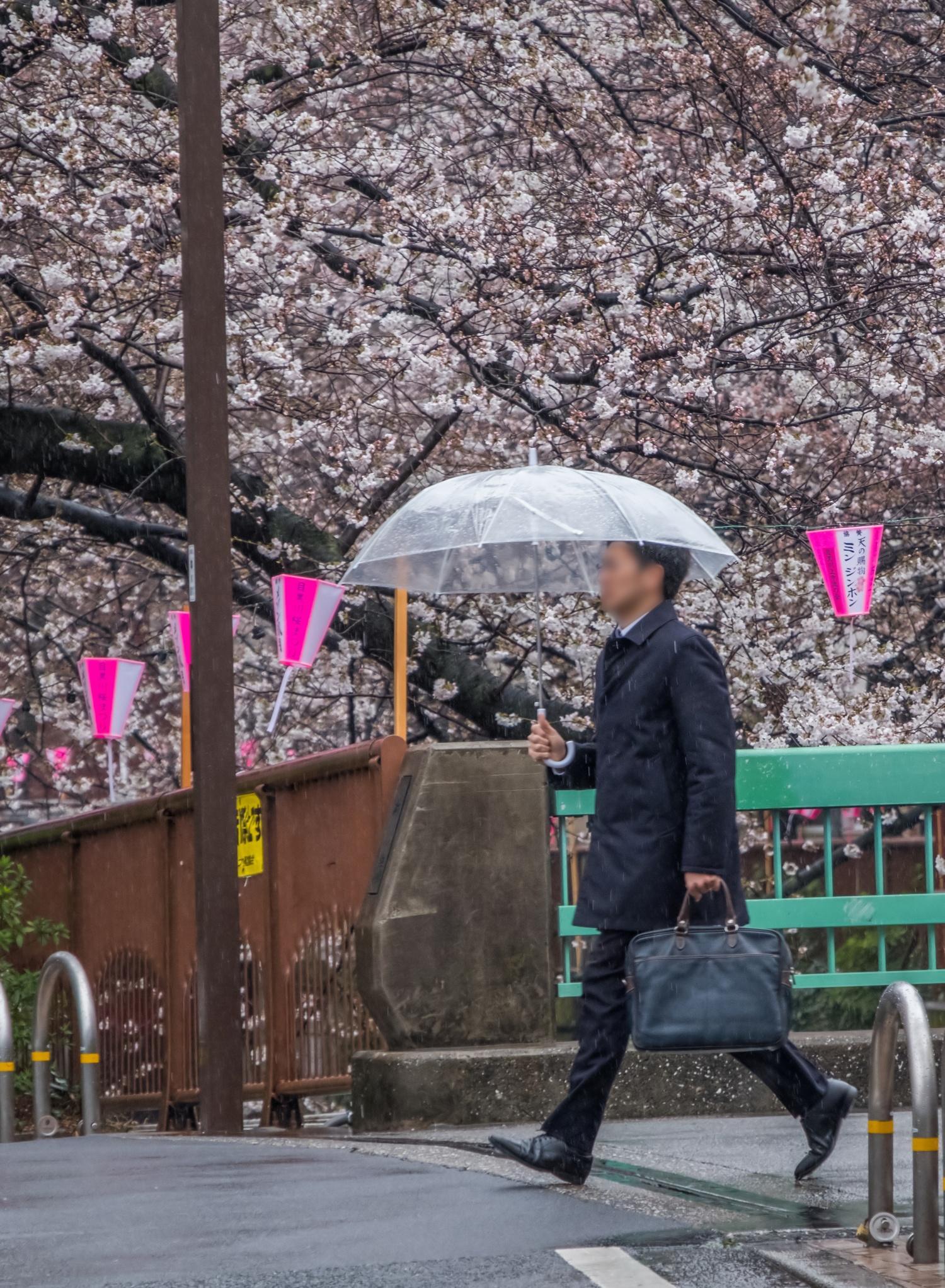}
\end{center}
\end{rolebubble}

\begin{rolebubble}{AssistantBg}{AssistantBorder}{LabelAssistant}{Assistant}
The image shows a serene outdoor scene with cherry blossoms in full bloom.
A person holding an umbrella walks across a bridge.
Pink lanterns with text are strung along the area, likely part of a festival.

I will zoom in to read the text on the pink lantern more clearly.

\smallskip
\begin{toolcall}
\{"name": "crop\_image\_normalized",\par
\ "arguments": \{"bbox\_2d": [0.0, 0.3, 0.4, 0.6], "target\_image": 1\}\}
\end{toolcall}
\end{rolebubble}

\begin{rolebubble}{UserBg}{UserBorder}{LabelUser}{Tool Result}
Here is the cropped image:

\medskip
\begin{center}
  \includegraphics[width=0.50\linewidth]{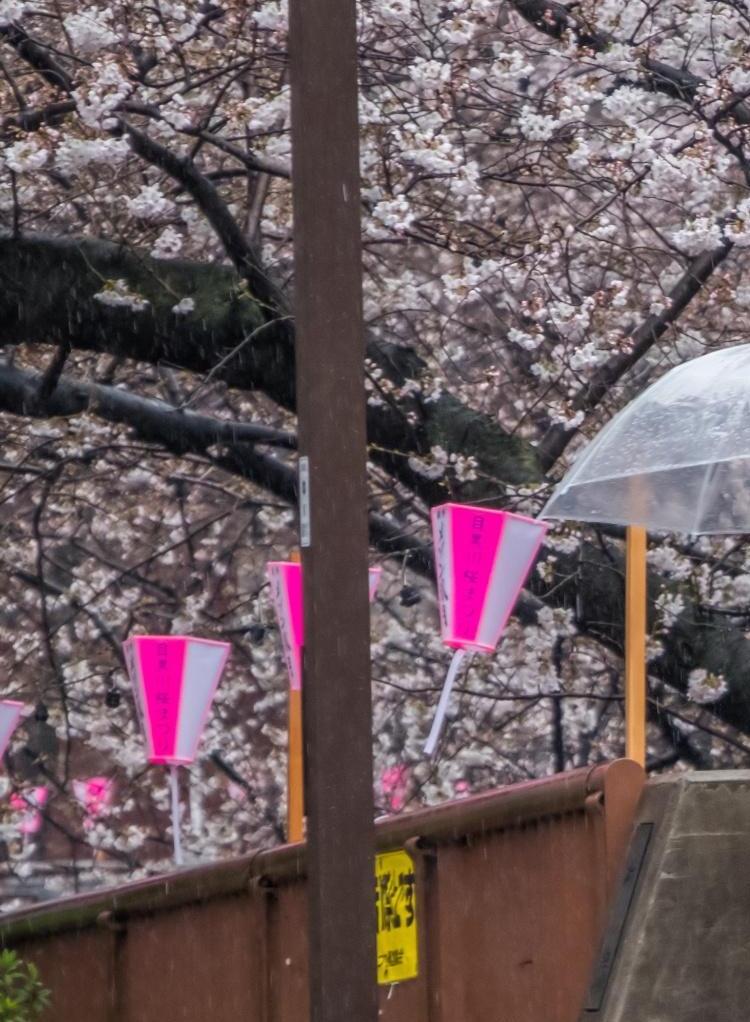}
\end{center}
\end{rolebubble}

\begin{rolebubble}{AssistantBg}{AssistantBorder}{LabelAssistant}{Assistant}
I can see the lanterns more clearly now, but the text needs a closer look.

\smallskip
\begin{toolcall}
\{"name": "crop\_image\_normalized",\par
\ "arguments": \{"bbox\_2d": [0.1, 0.1, 0.2, 0.2], "target\_image": 2\}\}
\end{toolcall}
\end{rolebubble}

\begin{rolebubble}{UserBg}{UserBorder}{LabelUser}{Tool Result}
Here is the cropped image:

\medskip
\begin{center}
  \includegraphics[width=0.35\linewidth]{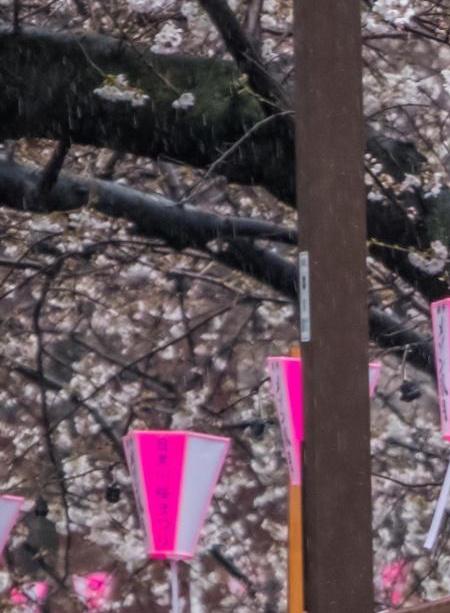}
\end{center}
\end{rolebubble}

\begin{rolebubble}{AssistantBg}{AssistantBorder}{LabelAssistant}{Assistant}
The close-up clearly shows the text on the pink lantern reads
\textbf{``Sakura Festival''}.

\[
  \{A\}
\]
\end{rolebubble}

\end{tcolorbox}
\begin{tcolorbox}[datasetbox, attach boxed title to top center={yshift=-2mm},
  title={Viscot (LVR) Example}]
\begin{rolebubble}{SystemBg}{SystemBorder}{LabelSystem}{System}
You are a helpful assistant.
\end{rolebubble}
\begin{rolebubble}{UserBg}{UserBorder}{LabelUser}{User}
What are the young men doing with their hands?
Provide a short and direct response.
\medskip
 
\begin{center}
  \includegraphics[width=0.65\linewidth]{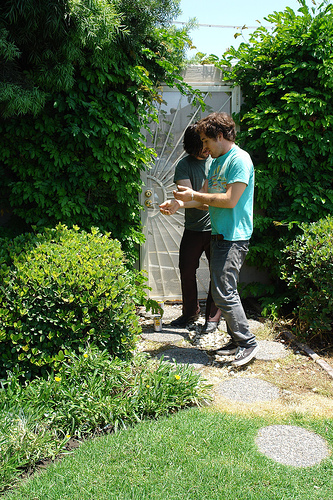}
\end{center}
\end{rolebubble}
\begin{rolebubble}{AssistantBg}{AssistantBorder}{LabelAssistant}{Assistant}
To determine what the young men are doing with their hands, I will focus on the hand region of the individuals visible in the image. I will draw a red bounding box around that region at coordinates \texttt{[0.482, 0.396, 0.548, 0.45]}.
\end{rolebubble}
\begin{rolebubble}{UserBg}{UserBorder}{LabelUser}{Tool Result}
\medskip
\begin{center}
  \includegraphics[width=0.50\linewidth]{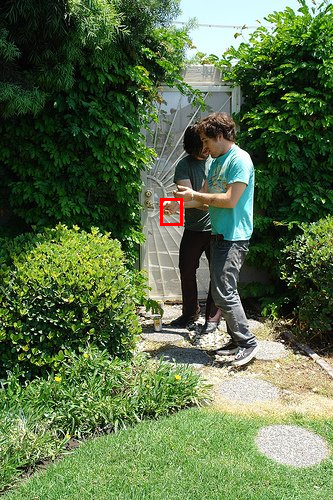}
\end{center}
\end{rolebubble}
\begin{rolebubble}{AssistantBg}{AssistantBorder}{LabelAssistant}{Assistant}
They are looking at their hands.
\end{rolebubble}
\end{tcolorbox}

\begin{tcolorbox}[datasetbox, attach boxed title to top center={yshift=-2mm},
  title={ThinkMorph Example: Chart Refocus}]

\begin{rolebubble}{SystemBg}{SystemBorder}{LabelSystem}{System}
You are a helpful assistant.
\end{rolebubble}

\begin{rolebubble}{UserBg}{UserBorder}{LabelUser}{User}
What was the total amount of towns, cities, and villages in the US in 2019 with population less than 25000?

\medskip\noindent

\medskip
\begin{center}
  \includegraphics[width=0.65\linewidth]{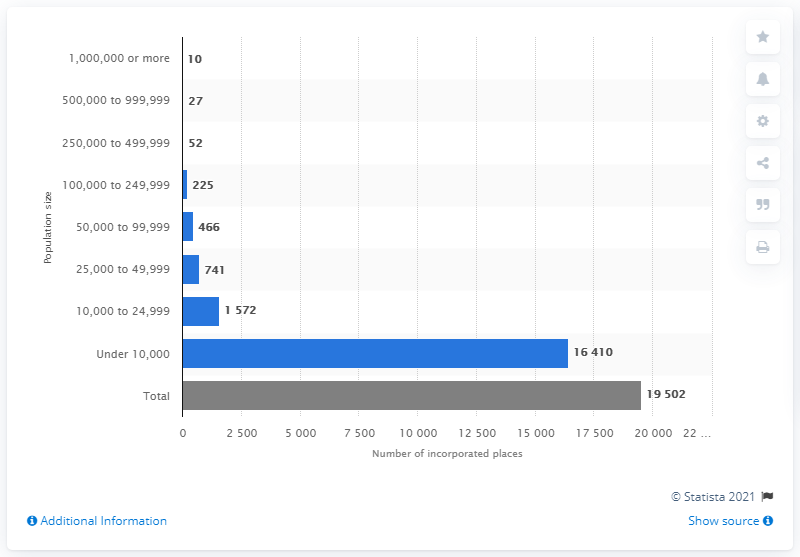}
\end{center}
\end{rolebubble}

\begin{rolebubble}{AssistantBg}{AssistantBorder}{LabelAssistant}{Assistant}
To answer the question about the total number of towns, cities, and villages in the US in 2019 with a population less than 25,000, I need to identify the relevant segments in the chart that correspond to this population range. The visualization displays the number of incorporated places grouped by population brackets. Specifically, I should look for the bars labeled 'Under 10,000' and '10,000 to 24,999', since both categories fall under the threshold of 25,000 residents. No other category applies, as the next smallest bracket, '25,000 to 49,999', exceeds the limit set by the query. By highlighting these two horizontal bars, I can clearly indicate the areas that collectively represent all places with populations below 25,000. This focus directly connects to the population constraint in the question and isolates the relevant data needed to compute the total. Therefore, the manipulation involves drawing a red bounding box around both the 'Under 10,000' and the '10,000 to 24,999' rows in the chart.
\end{rolebubble}

\begin{rolebubble}{UserBg}{UserBorder}{LabelUser}{Tool Result}

\medskip
\begin{center}
  \includegraphics[width=0.50\linewidth]{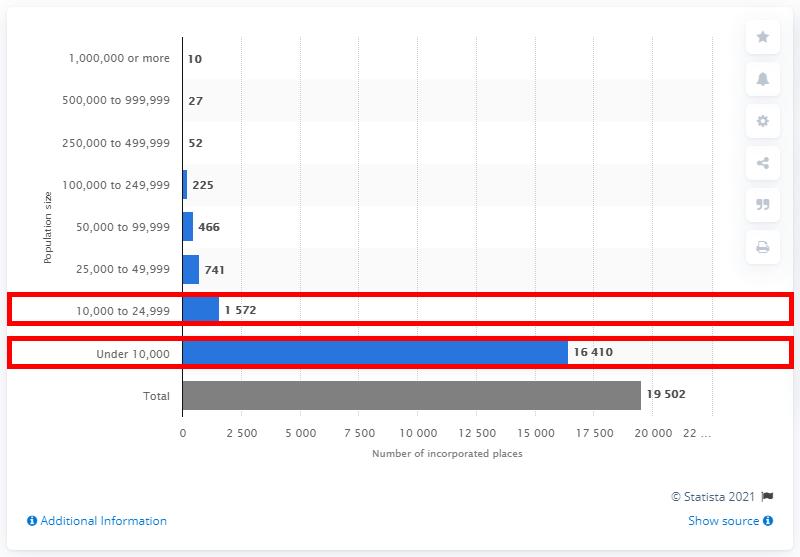}
\end{center}
\end{rolebubble}

\begin{rolebubble}{AssistantBg}{AssistantBorder}{LabelAssistant}{Assistant}
To determine the total number of towns, cities, and villages in the US in 2019 with a population less than 25,000, I first identify the relevant population categories in the chart. The two categories that fit this criterion are 'Under 10,000' and '10,000 to 24,999', which are highlighted with red boxes in the edited visualization. The numbers associated with these categories are 16,410 for 'Under 10,000' and 1,572 for '10,000 to 24,999'. Summing these two values gives the total number of incorporated places with populations below 25,000: 16,410 + 1,572 = 17,982. Therefore, the total amount of towns, cities, and villages in the US in 2019 with population less than 25,000 is 17,982.
\end{rolebubble}
\end{tcolorbox}

\begin{tcolorbox}[datasetbox, attach boxed title to top center={yshift=-2mm},
  title={ThinkMorph Example: Visual Search}]

\begin{rolebubble}{SystemBg}{SystemBorder}{LabelSystem}{System}
You are a helpful assistant.
\end{rolebubble}

\begin{rolebubble}{UserBg}{UserBorder}{LabelUser}{User}
What is the fruit that is on the plate called?


\medskip\noindent

\medskip
\begin{center}
  \includegraphics[width=0.65\linewidth]{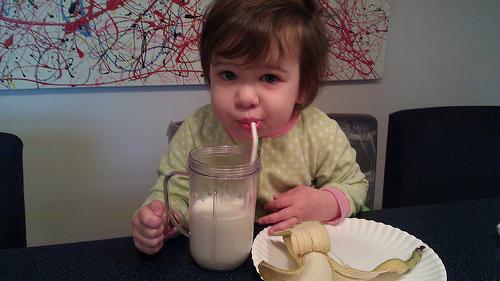}
\end{center}
\end{rolebubble}

\begin{rolebubble}{AssistantBg}{AssistantBorder}{LabelAssistant}{Assistant}
To answer the question about the fruit on the plate, begin by visually scanning the image for a plate. The plate is located in the lower right portion of the image, on the tabletop in front of the child. Next, look closely at what is placed on the plate. There is a pale yellow, curved object with a thick peel partially removed, exposing the inner fruit. The shape, color, and partially peeled nature are strong visual cues that this is a fruit. The red bounding box should be drawn around this item on the plate to highlight it as the focal point for identifying the fruit.
\end{rolebubble}

\begin{rolebubble}{UserBg}{UserBorder}{LabelUser}{Tool Result}

\medskip
\begin{center}
  \includegraphics[width=0.50\linewidth]{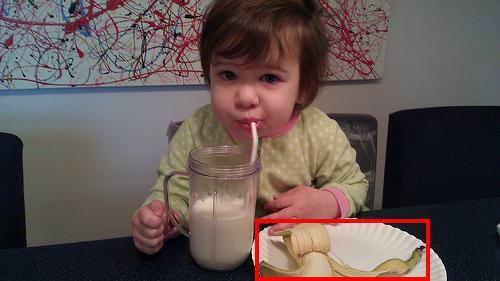}
\end{center}
\end{rolebubble}

\begin{rolebubble}{AssistantBg}{AssistantBorder}{LabelAssistant}{Assistant}
With the fruit on the plate now highlighted by the red bounding box, it's easy to closely examine its characteristics. The fruit has a distinctive elongated, curved shape and a pale yellow peel that is partially removed, revealing the soft, creamy flesh inside. These visual details—such as the color, the way the peel splits away from the fruit, and the overall shape—are classic identifiers for a banana. By focusing on the region inside the red bounding box, we can confidently determine that the fruit on the plate is a banana.
\end{rolebubble}
\end{tcolorbox}

\begin{tcolorbox}[datasetbox, attach boxed title to top center={yshift=-2mm},
  title={ThinkMorph Example: Spatial Navigation}]

\begin{rolebubble}{SystemBg}{SystemBorder}{LabelSystem}{System}
You are a maze solver. Your goal is to guide a player from the start to the goal on a grid map while avoiding holes. The player can move one square at a time in the directions left (L), right (R), up (U), or down (D). The frozen lake is not slippery; the player will always move in the intended direction. Moving off the edge or falling into a hole results in failure. Reaching the goal means success.
\end{rolebubble}

\begin{rolebubble}{UserBg}{UserBorder}{LabelUser}{User}
Provide your solution as a sequence of moves wrapped in \texttt{boxed}\{\}, such as \{L,R,U,D\}. The moves should be comma-separated.


\medskip\noindent

\medskip
\begin{center}
  \includegraphics[width=0.65\linewidth]{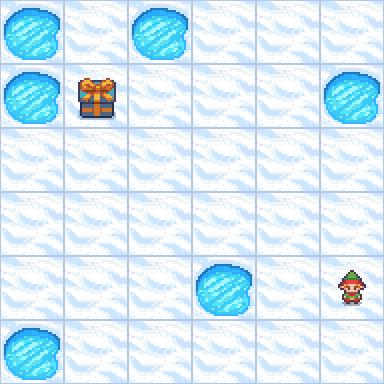}
\end{center}
\end{rolebubble}

\begin{rolebubble}{AssistantBg}{AssistantBorder}{LabelAssistant}{Assistant}
To get a clear sense of the challenge, I should first lay out the details of the maze. The grid is 6 rows by 6 columns. The starting point is in the bottom right corner, specifically at row 5, column 5 if counting from the top-left as (0,0). The goal is near the top left, at row 1, column 1.

There are several ice holes scattered around the grid:
\begin{itemize}
    \item 
\item The first row has holes at positions (0,0) and (0,2).
\item The second row contains two holes: one at (1,0) and another at (1,5).
\item The fourth row has a hole at (4,3).
\item The last row (row 5) has a hole at (5,0).
\end{itemize}

All other squares are safe frozen surfaces. Time to move on to plotting the solution and finding the optimal path.
\end{rolebubble}

\begin{rolebubble}{UserBg}{UserBorder}{LabelUser}{Tool Result}

\medskip
\begin{center}
  \includegraphics[width=0.50\linewidth]{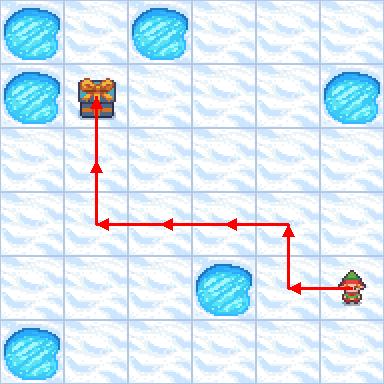}
\end{center}
\end{rolebubble}

\begin{rolebubble}{AssistantBg}{AssistantBorder}{LabelAssistant}{Assistant}
Reviewing my solution, the red path begins at the starting square in the bottom right and follows the moves: Left, Up, Left, Left, Left, Up, Up.

Step by step:
\begin{itemize}
\item First, move left to avoid the edge.
\item Then, up to stay clear of the hole below.
\item Continue left three times, traversing safe frozen squares and bypassing the hole above.
\item Go up twice to reach the goal square near the top left, successfully avoiding all holes and edges.

\end{itemize}
Each move lands on a safe frozen surface, and the path finishes directly on the goal. This confirms the route is valid and optimal.

Final solution: \{L,U,L,L,L,U,U\}

\end{rolebubble}
\end{tcolorbox}

\subsection{Evaluation Data}
We draw our evaluation tasks from MMVP \citep{MMVP}, V* \citep{vstar},
and subsets of the Blink dataset
\citep{Blink}. Table~\ref{tab:eval_datasets} provides a brief
description of each dataset along with the number of evaluation
samples.  

\begin{table}[ht]
    \centering
    \begin{tabular}{llc}
        \toprule
        \textbf{Dataset} & \textbf{Description} & \textbf{No. Samples} \\
        \midrule
        V* & Object attributes & 191 \\
        MMVP & Perception Robustness & 300 \\
        Counting & Counting objects & 120 \\
        IQ & Pattern Matching & 150 \\
        Jigsaw & Multi-image jigsaw resolution & 150 \\
        Relative Reflectance & Perception  & 134 \\
        Spatial Relation & Relation between objects & 143  \\
        \bottomrule
    \end{tabular}
     \caption{Overview of datasets and number of evaluation samples.}
    \label{tab:eval_datasets}
\end{table}

\section{Limitations and Future Work}
\label{sec:limitations}

While \textsc{Pearl} demonstrates strong performance across three
training regimes and multiple model scales, several limitations remain.

\textbf{Planning over Learned Embeddings.}
\textsc{Pearl} internalizes the \emph{effects} of tool use in the latent
space but does not explicitly plan sequences of actions at inference.
The predictive embeddings learned by $\mathcal{L}_{\text{JEPA}}$ encode
a holistic representation of the full expert trajectory, which implicitly
captures planning structure, but the model does not reason step-by-step
over these representations at test time. A natural extension would be to
use the learned embeddings as a latent world model for explicit multi-step
planning, enabling the model to reason about longer action horizons
without invoking external tools.

\textbf{Interpretability.}
Because \textsc{Pearl} operates entirely in a continuous embedding space,
the learned trajectory representations are not directly interpretable.
Unlike reconstruction-based methods, which at least nominally produce
latent tokens aligned with intermediate image edits, \textsc{Pearl} makes
no claim about what individual dimensions of the embedding encode. While
the t-SNE visualizations in Figure~\ref{fig:vizplot} confirm that
the representations are structured and task-discriminative, understanding
\emph{what} the model has internalized about tool use remains an open
question. Developing probing methods or disentangled representations that
make the learned latent structure more transparent is an important
direction for future work.

\textbf{Training Data Coverage.}
Our experiments are limited to settings for which open-source expert
trajectory data exists. In particular, we do not evaluate the
multiple-type, multiple tool call setting due to the absence of suitable
training data. As more diverse trajectory datasets become available, we
expect \textsc{Pearl}'s trajectory-level embedding objective to generalize
naturally to richer tool-use settings, given that it places no constraints
on the number or type of tools present in $\mathcal{R}$.

\textbf{Training Cost.}
\textsc{Pearl} requires two separate forward passes per training example
to encode the input view and the trajectory view independently, which
roughly doubles the training-time compute relative to standard SFT.
While this overhead does not affect inference, it may be prohibitive at
very large scale. Exploring more efficient encoding strategies, such as
shared encoders with cross-view masking or cached trajectory embeddings, is a promising avenue for reducing this cost.

\end{document}